\documentclass[10pt,twocolumn,letterpaper]{article}

\usepackage{cvpr}              

\usepackage[T1]{fontenc}
\usepackage{graphicx}
\usepackage{amsmath}
\usepackage{amssymb}
\usepackage{booktabs}
\usepackage{multirow}
\usepackage{bm}
\usepackage[table]{xcolor}
\usepackage{makecell}                 
\usepackage{booktabs}                 
\usepackage[ruled,vlined]{algorithm2e}
\usepackage{threeparttable}

\usepackage[pagebackref,breaklinks,colorlinks]{hyperref}
\usepackage{url}

\usepackage[capitalize]{cleveref}
\crefname{section}{Sec.}{Secs.}
\Crefname{section}{Section}{Sections}
\Crefname{table}{Table}{Tables}
\crefname{table}{Tab.}{Tabs.}

\begin{document}

\title{Efficient Generalization Improvement Guided by Random Weight Perturbation}

\author{
Tao Li, Weihao Yan, Zehao Lei, Yingwen Wu, Kun Fang, Ming Yang, Xiaolin Huang
\and
Department of Automation, Shanghai Jiao Tong University
\\
{\tt\small \{li.tao, ywh926934426, lzhsjstudy, yingwen\_wu, fanghenshao, mingyang, xiaolinhuang\}@sjtu.edu.cn}
}

\maketitle

\begin{abstract}
    To fully uncover the great potential of deep neural networks (DNNs), various learning algorithms have been developed to improve the model's generalization ability. Recently, sharpness-aware minimization (SAM) establishes a generic scheme for generalization improvements by minimizing the sharpness measure within a small neighborhood and achieves state-of-the-art performance. However, SAM requires two consecutive gradient evaluations for solving the min-max problem and inevitably doubles the training time. In this paper, we resort to filter-wise random weight perturbations (RWP) to decouple the nested gradients in SAM. 
    Different from the small adversarial perturbations in SAM, RWP is softer and allows a much larger magnitude of perturbations.
    Specifically, we jointly optimize the loss function with random perturbations and the original loss function: the former guides the network towards a wider flat region while the latter helps recover the necessary local information. These two loss terms are complementary to each other and mutually independent. Hence, the corresponding gradients can be efficiently computed in parallel, enabling nearly the same training speed as regular training. As a result, we achieve very competitive performance on CIFAR and remarkably better performance on ImageNet (e.g. $\mathbf{ +1.1\%}$) compared with SAM, but always require half of the training time. 
    The code is released at \url{https://github.com/nblt/RWP}.


\end{abstract}

\section{Introduction}
\label{sec:intro}

Modern deep neural networks (DNNs) are typically over-parameterized and contain millions or even billions of parameters. Such great model capability of DNNs provides a huge functional 
space and is capable of achieving 
state-of-the-art 
performance  across a variety of tasks \cite{tan2019efficientnet, kolesnikov2020big, liu2021swin, radford2021learning}.
Since the number of parameters completely exceeds that of samples, DNNs could easily memorize the training data and overfit them eventually, even with random labels \cite{zhang2021understanding}. 
Thus, it is crucial to develop effective training algorithms that  guide the network to attain fantastic interpolation 
and  generalize well beyond the training set \cite{neyshabur2017exploring}.

\begin{figure}[!t]
    \centering
    \includegraphics[width=0.9\linewidth]{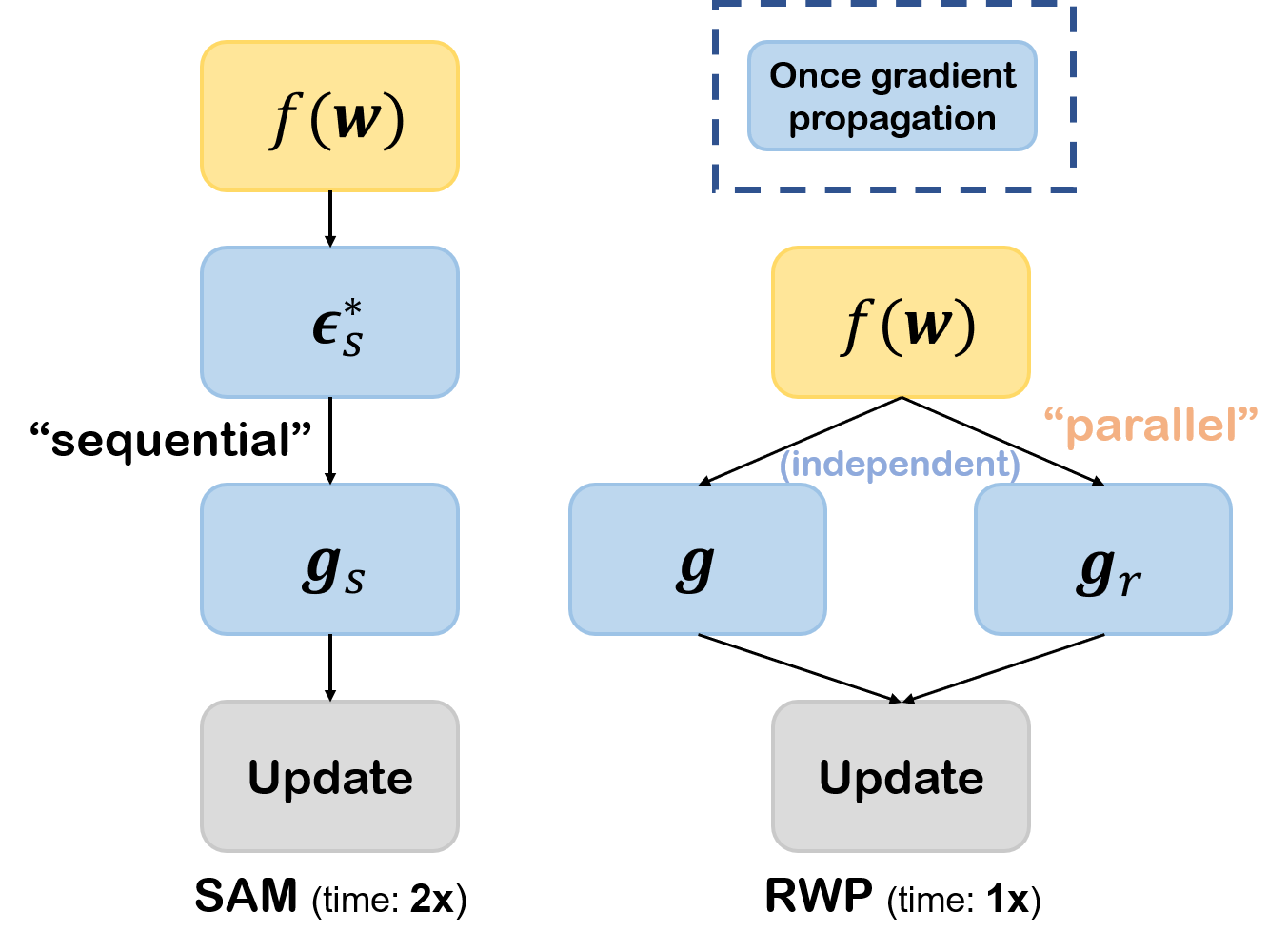}
    \caption{Illustration of SAM and RMP. SAM involves two sequential gradient evaluations, hence requiring {$\bm 2 \boldsymbol{\times}$} training time as regular training. For RWP, the two gradient steps are independent of each other and can be perfectly parallelized, enabling nearly the same training speed ($\bm {1} \boldsymbol{\times}$) as regular training.}
    \label{fig:comparison}
\end{figure} 

Many works are devoted to improving the generalization of DNNs \cite{szegedy2016rethinking, izmailov2018averaging, zhang2018mixup, zhang2019lookahead, foret2020sharpness}. 
A widely accepted premise is that flat minima will adapt better to the potential distribution shift of data and possess better generalization \cite{hochreiter1997flat,dinh2017sharp, li2018visualizing}.
Based on this, Foret et al. \cite{foret2020sharpness} proposed sharpness-aware minimization (SAM), which provides a generic scheme for improving generalization and achieves state-of-the-art performance among various image classification tasks.
Specifically, SAM formulates the optimization target as a min-max problem and tries to minimize the training loss under worst-case weight perturbations.
Such worst loss in a neighborhood (defined by a norm ball centered at the current weights $\boldsymbol{w}$) is referred to as ``sharpness'', by minimizing which SAM could attain much flatter minima.

In each iteration,
SAM has to solve an inner max problem to obtain the worst-case perturbation $\bm{\epsilon}_s^*$ 
and then performs gradient descent step with the gradient at the perturbed weights $\boldsymbol{w}+\bm{\epsilon}_s^*$. Thus, unfortunately, the training speed of SAM is $2\times$ slower than that of regular training due to the consecutive gradient steps involved, making it prohibitively time-consuming.

In this paper, we propose to adopt random weight perturbations $\bm{\epsilon}_r$ rather than the worst-case perturbations $\bm{\epsilon}_s^*$ to retain training efficiency and meanwhile achieve good generalization improvements.
As the perturbations are generated purely based on the statistical distribution of parameters, the successive gradient computations in SAM are no longer required. 
Random perturbations are weaker in threatening the model than adversarial ones under the same perturbation magnitude. 
This weakening allows a much larger perturbation magnitude, with which SAM's optimization may crash for the substantially increased difficulty.
Using such random perturbation, we will obtain a wider flat region, and thus good generalization improvements could be expected. 

\begin{figure}[!t]
 \centering
 \begin{subfigure}{0.495\linewidth}
 \centering
  \includegraphics[width=1\linewidth]{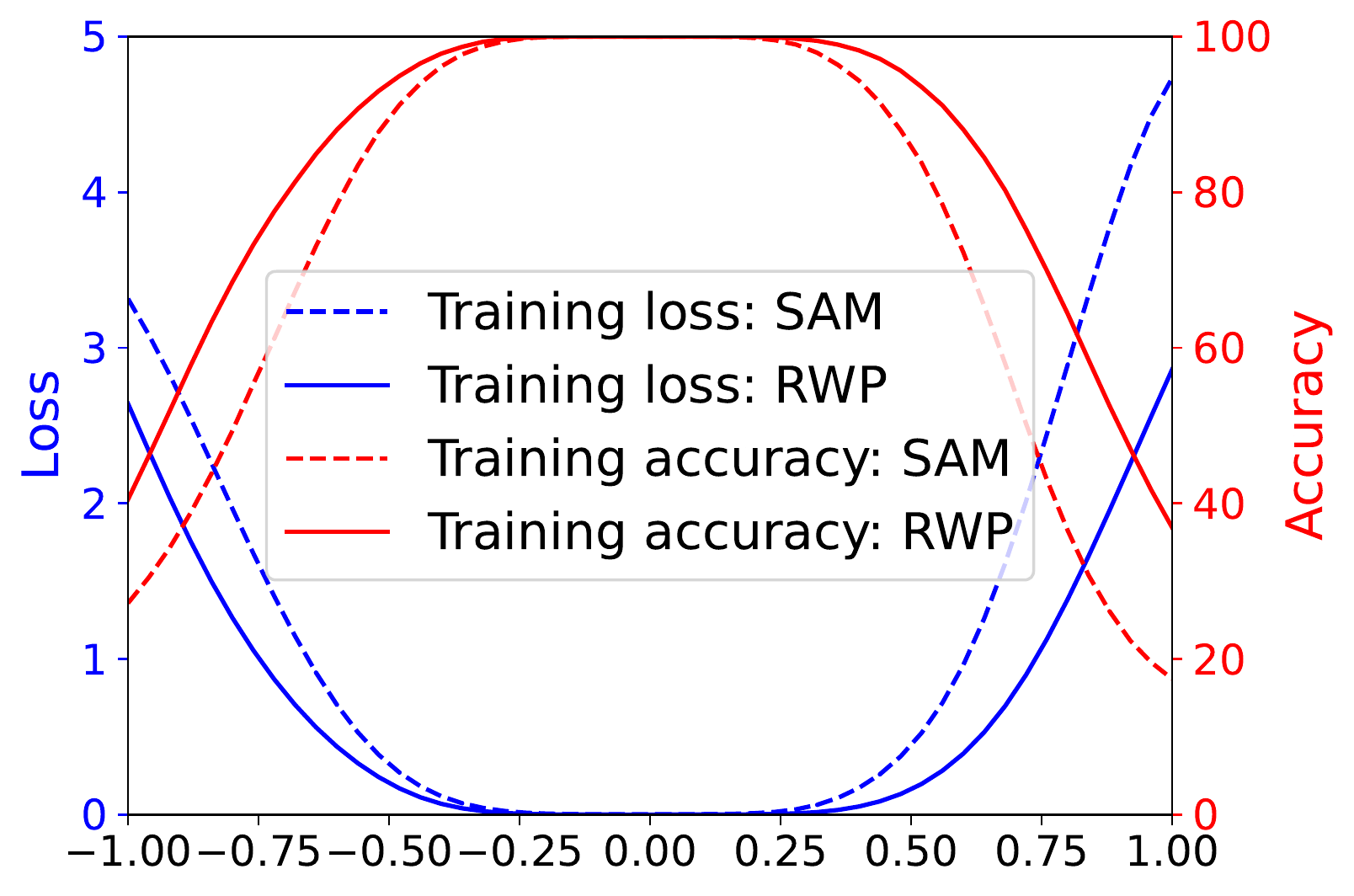}
 \end{subfigure}
 \begin{subfigure}{0.495\linewidth}
 \centering
  \includegraphics[width=1\linewidth]{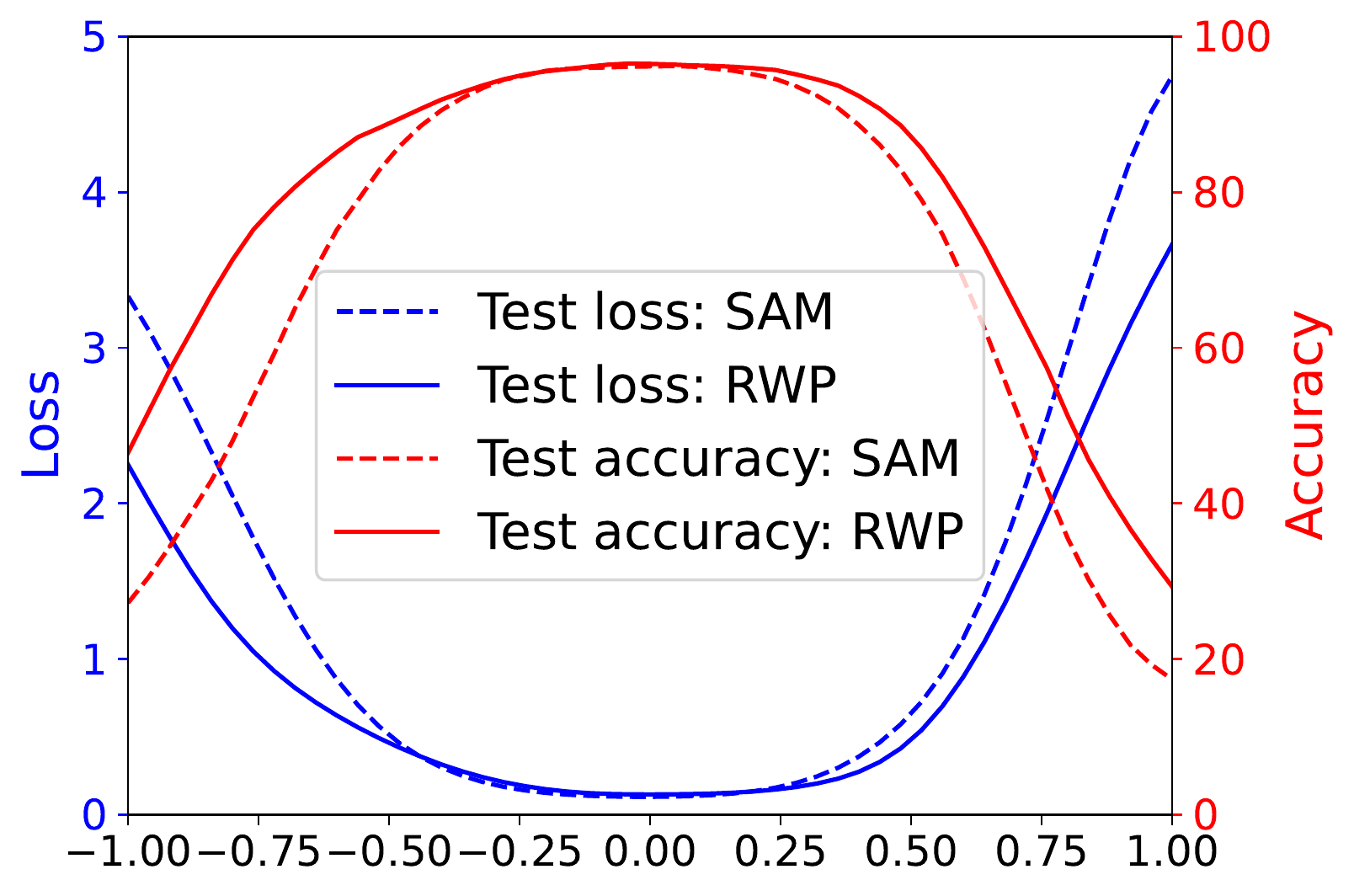}
 \end{subfigure}
 \caption{ 
 Landscape visualizations of the empirical loss obtained from the ResNet-18 models trained with SAM (dotted) and RWP (solid) on CIFAR-10. 
 \textbf{Left}: on training set; \textbf{right}: on test set.
 We plot the loss/accuracy curves along a normalized random direction,
 following the plotting technique proposed in \cite{li2018visualizing}. 
 RWP could obtain a wider flat region than SAM.
}
 \label{fig:visualization}
\end{figure}

Imposing random perturbations, rather than adversarial ones, could save computational time and meanwhile keep good generalization. But when applying the gradient at the perturbed point to the current solution, the information is ``blurred'' due to the large perturbation magnitude. 
To resolve this problem, 
we propose to jointly optimize the original loss function as well as the perturbed loss function. 
The two objectives are complementary: 
the gradient of the perturbed loss provides a broader view of the landscape that helps escape from a local optimum, while the gradient of the original loss pays attention to the local and recovers the original loss information necessary for an optimal solution.
Through the close cooperation of these two gradients, we could sufficiently dive into the low-loss region of DNNs' landscape meanwhile staying away from the local optimum that is sub-optimal on a larger scale.

The above two losses are separable and the corresponding gradients 
can be efficiently computed \emph{in parallel},  
which may halve the training time from SAM, where serial calculation on two gradients is inevitable.
\cref{fig:comparison} illustrates the difference between SAM and our approach, dubbed \emph{\textbf{R}andom \textbf{W}eight \textbf{P}urturbation (RWP)}.
As a result, we obtains a minimum in a wider flat region (\cref{fig:visualization}) and achieve superior performance (\cref{tab:cifar}, \cref{tab:cifar100}, \cref{tab:ImageNet}), compared with SAM, e.g., up to 1.1\% accuracy improvement on ImageNet \cite{deng2009imagenet}, but always consuming only \emph{half} of training time.

Our main contributions are as follows:
\begin{itemize}
    \item We propose a simple training algorithm that attains significant generalization improvements by jointly optimizing the original and the perturbed loss functions.
    \item Compared to SAM, our method is easy to parallelize, allowing nearly twice the training speed while enjoying a comparable or even better performance.
    \item We conduct extensive experiments with various architectures on benchmark image classification tasks to demonstrate the efficiency and superior performance of our methods, especially on large-scale datasets.
\end{itemize}


\section{Related Works}
\label{sec:relatedworks}

\paragraph{Sharp Minima and Generalization.}
The connection between the flatness of local minima and generalization has received a rich body of studies \cite{hochreiter1997flat, chaudhari2017entropy, keskar2017large,dinh2017sharp, izmailov2018averaging, li2018visualizing}.
Recently, many works try to improve the model generalization by approaching flatter minima \cite{chaudhari2017entropy, wen2018smoothout, tsuzuku2020normalized, foret2020sharpness, zheng2021regularizing, bisla2022low}. For example, \cite{chaudhari2017entropy} proposed Entropy-SGD that actively searches for flat regions by minimizing local entropy.
\cite{wen2018smoothout} proposed SmoothOut framework to smooth out the sharp minima and obtain generalization improvements.
Notably, sharpness-aware minimization (SAM) \cite{foret2020sharpness} provided a generic training scheme for seeking flat minima by formulating a min-max problem and encouraging parameters sitting in neighborhoods with uniformly low loss. It achieves state-of-the-art generalization improvements on image classification tasks. 
Later, a line of works improves the SAM's performance from the perspective of the neighborhood's geometric measure \cite{kwon2021asam, kim2022fisher} or surrogate loss function \cite{zhuang2022surrogate}.

\paragraph{Efficient Sharpness-aware Minimization.}
As SAM requires two consecutive gradient propagation for solving the min-max problem,
its training speed is $2\times$ slower than that of regular SGD training. Many methods have been developed to improve the training efficiency \cite{du2022efficient, liu2022towards, du2022sharpness}. For example, \cite{du2022efficient} proposed ESAM that adopts two acceleration strategies for SAM's two updating steps and achieves up to 30\% computational saving. \cite{liu2022towards} proposed LookSAM, which periodically calculates the inner gradient ascent step to reduce the additional computation. 
\cite{mi2022make} accelerates the SAM's training by introducing a sparsified binary mask for perturbations.
\cite{du2022sharpness} proposed a novel trajectory loss to replace the gradient accent step of SAM and achieve almost ``free'' sharpness-aware minimization. 
Different from previous approaches, we improve the training efficiency of SAM by parallelizing the two consecutive gradient steps in each iteration, allowing almost the same training speed as regular training with comparable or even better generalization.


\paragraph{Regularization Methods.} 
Various regularization methods play a fundamental role in generalization improvements.
Apart from the well-known regularization techniques such as weight decay \cite{krogh1991simple}, dropout \cite{srivastava2014dropout}, label smoothing \cite{szegedy2016rethinking} and normalization \cite{ioffe2015batch, ba2016layer}, MixUp \cite{zhang2018mixup} encourages the network to behave linearly across different samples. 
Flooding \cite{ishida2020we} directly forces the training loss to stay above zero to avoid meaningless overfitting. 
\cite{zhao2022penalizing} penalized the gradient norm of loss function during optimization to efficiently improve the generalization.
Data augmentation techniques, such as Cutout \cite{devries2017improved}, CutMix \cite{yun2019cutmix}, RandAugment \cite{cubuk2020randaugment} and AutoAugment \cite{cubuk2019autoaugment}, also regularize deep networks from overfitting by diversifying the training data, thereby improving the performance.
We impose regularization by requiring the network to perform well under random weight perturbations. Our method is orthogonal to previous regularization methods, and it is possible to combine them to attain further generalization improvements.

\section{Method}
\label{sec:method}
In this section, we will introduce our RWP algorithm in detail. We first elaborate on how to generate random perturbations that effectively perturb the model. Then we formulate our optimization target and describe our training procedure. Finally, we analyze the filter norm distribution of the models trained by different methods.



\subsection{Preliminaries}
Let $f(\boldsymbol{x};\boldsymbol{w})$ be the neural network function with trainable parameters $\boldsymbol{w}$. The loss function over a pair of data point $(\boldsymbol{x},\boldsymbol{y})$ is denoted as  $L(f(\boldsymbol{x};\boldsymbol{w}), \boldsymbol{y})$. Given the datasets $\mathcal{S}=\{ (\boldsymbol{x}_i,\boldsymbol{y}_i) \}_{i=1}^n$ drawn from data distribution $\mathcal{D}$ with i.i.d. condition, the empirical loss can be defined as $L_{\mathcal{S}}(\boldsymbol{w})=\frac{1}{n}\sum_{i=1}^n L(f(\boldsymbol{x}_i;\boldsymbol{w}), \boldsymbol{y}_i)$. 

In the SAM algorithm \cite{foret2020sharpness}, 
a flat minimum is sought
by minimizing the worst-case loss in a neighborhood (define as a norm ball), i.e.
\begin{equation}
    L^{\rm SAM}_{\mathcal{S}}(\boldsymbol{w})= \max_{\| \bm{\epsilon}_s \|_2 \le \rho} L_{\mathcal{S}}(\boldsymbol{w}+\bm{\epsilon}_s),
    \label{equ:sam}
\end{equation}
where $\rho$ is the radius of the ball. To optimize $L^{\rm SAM}_{\mathcal{S}}$, we first have to find the worst-case perturbations $\bm{\epsilon}_s^*$ for the inner max problem. In practice, \cite{foret2020sharpness}
approximates \cref{equ:sam} via first-order approximations and obtains:
\begin{equation}
    \bm{\epsilon}_s^*\approx \mathop{ \arg \max}\limits_{\| \bm{\epsilon}_s \|_2\le \rho} \bm{\epsilon}_s ^{\top} \nabla_{\boldsymbol{w}} L_{\mathcal{S}}(\boldsymbol{w})\approx \rho \frac{\nabla_ {\boldsymbol{w}} L_{\mathcal{S}}(\boldsymbol{w})}{\| \nabla_{\boldsymbol{w}} L_{\mathcal{S}}(\boldsymbol{w}) \|_2}.
    \label{equ:first-order}
\end{equation}
Then the gradient at the perturbed weight  $\boldsymbol{w}+\bm{\epsilon^*_s}$ is computed for updating the model:
\begin{equation}
    \nabla L^{\rm SAM}_{\mathcal{S}} ({\boldsymbol{w}}) \approx
    \nabla L_{\mathcal{S}} ({\boldsymbol{w}}) |_{\boldsymbol{w}+\bm{\epsilon}_s^*}.
\end{equation}
Hence, the training speed of SAM is strictly $2\times$ slower than that of regular SGD training as it involves two sequential gradient calculations for each iteration.

\begin{figure}[!t]
    \centering
    \includegraphics[width=0.85\linewidth]{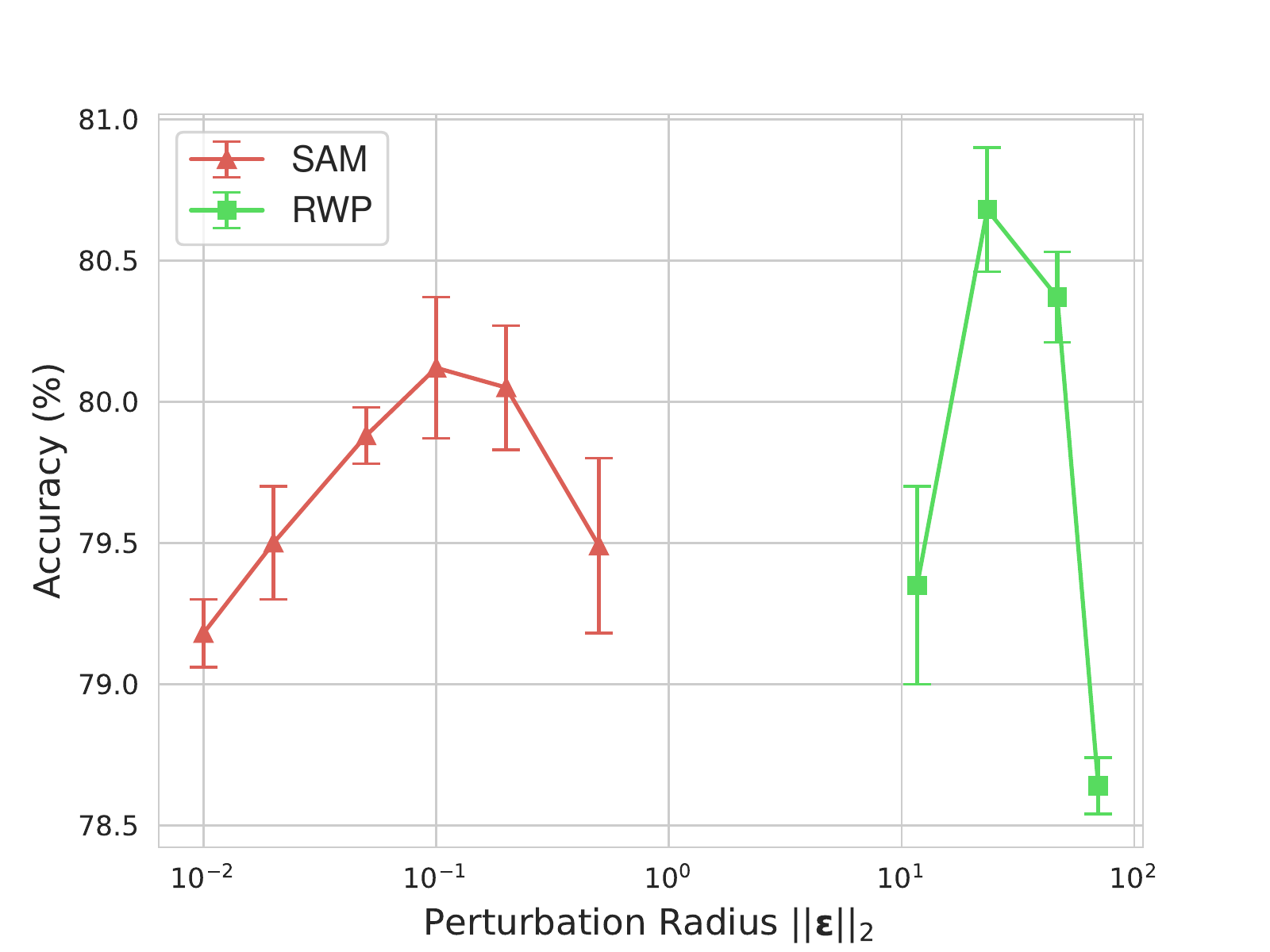}
    \caption{
    Comparisons on the different perturbation radii of SAM and RWP.
    The experiments are conducted on CIFAR-100 with ResNet-18 model. We examine SAM with perturbation radius $\rho$ in $\{ 0.01, 0.02, 0.05, 0.10, 0.20, 0.50 \}$ and RWP with $\gamma$ in $\{ 0.005, 0.01, 0.02, 0.03 \}$.
    The radius of random perturbations in RWP is determined by averaging $1000$ times sampled at initialization.
    Note that abscissa is logarithmic and the perturbation radius of RWP is orders of magnitude larger than that of SAM. The weight norm $\|\boldsymbol{w}\|_2$ at initialization is $80.19$, far greater than $\rho$.
    }
    \label{fig:radius}
\end{figure} 

\subsection{Random Perturbation Generation}
SAM perturbs the model parameters along the direction of gradient accent, which precisely captures the sharpness in a linearized local neighborhood. However, the radius of such a neighborhood $\rho$ perceived by SAM could be very small: we typically set $\rho$ (the norm of perturbations) as 0.05 or 0.10 \cite{foret2020sharpness}, far smaller than the norm of model parameters $\| \boldsymbol{w} \|_2$ (which is as large as 80 for ResNet-18 as shown in \cref{fig:radius}).
When $\rho$ becomes a bit larger, the optimization difficulty of \cref{equ:sam} substantially increases due to the strict constraint of the inner maximum problem, resulting in poor representation learning as well as severe degrading of performance. 
Besides, the first-order approximation in \cref{equ:first-order} can no longer hold with the increase of $\rho$, and more steps of iterations for a precise $\bm{\epsilon}_s^*$ are expected. Crucially, finding the gradient accent direction inevitably increases the training time for each iteration, which could be unbearably time-consuming for large-scale problems.

Adopting random perturbations could be a good alternative since we no longer require the additional gradient accent step. Meanwhile, the perturbations are much ``softer'' and significantly larger perturbations are allowed, potentially enabling a broader view of the landscape. Actually, 
random perturbations have been tried in some previous works \cite{zheng2021regularizing} but attain minor generalization improvements. The failure is perhaps due to 1) a too-small magnitude of the perturbations and 2) ignorance of the network's structure. In this paper, we will try large random perturbations and take into account the structure of networks filter-wisely. In \cref{fig:radius}, we demonstrate that the perturbation radius required for RWP is orders of magnitude larger than that of SAM, and a too-small perturbation magnitude will severely degrade the performance.


\paragraph{Filter-wise noise generation.} 
Modern DNNs are highly scale-invariant \cite{dinh2017sharp}. The loss function does not change with parameter scaling when ReLU-nonlinearities and batch normalization \cite{ioffe2015batch} are applied. Hence, it is important to consider the filter-wise structure for  generating effective perturbations.
Following \cite{bisla2022low}, we suppose the model parameters $\boldsymbol{w}$ could be divided into $k$ filters, i.e., $\boldsymbol{w}=[\boldsymbol{w}_1,\boldsymbol{w}_2,\cdots,\boldsymbol{w}_k]$. Here, a ``filter'' represents the parameters that contribute to a single feature map (or a single neuron for fully connected layers). Then for each filter $\boldsymbol{w}_k$, we generate the corresponding perturbations from the Gaussian distribution $\mathcal{N}(0, \gamma \| \boldsymbol{w}_k \|_2)$ elementally, where $\gamma$ is a hyper-parameter that controls the magnitude of perturbations. Note that the filters with larger norm will receive stronger perturbations.

\subsection{Optimization Objective of RWP}
To attain similar perturbation strength as adversarial perturbations in SAM, random perturbations require a significantly larger magnitude, which can conversely impair the loss function and hurt the performance. To resolve such contradiction, we propose to jointly optimize the original loss  $L_{\mathcal{S}}(\boldsymbol{w})$ and perturbed loss  $L_{\mathcal{S}}(\boldsymbol{w+\bm{\epsilon}_r})$, i.e.,
\begin{equation}
    L^{\rm RWP}_{\mathcal{S}}(\boldsymbol{w}) = \alpha L_{\mathcal{S}}(\boldsymbol{w}) + (1 - \alpha) L_{\mathcal{S}}(\boldsymbol{w}+\bm{\epsilon}_r),
    \label{equ:loss}
\end{equation}
where $\alpha$ is a pre-given balance coefficient.
By mixing the original loss $L_{\mathcal{S}}(\boldsymbol{w})$ and ``blurred'' loss $L_{\mathcal{S}}(\boldsymbol{w+\bm{\epsilon}_r})$, we could obtain an adaptively smoothed loss function: the gradient of perturbed loss $\nabla L_{\mathcal{S}}(\boldsymbol{w}+\bm{\epsilon}_r)$ can hold a broad view of the current landscape around $\boldsymbol{w}$ and guide the network towards a flatter minimum, while the gradient of original loss $\nabla L_{\mathcal{S}}(\boldsymbol{w})$ recovers the necessary information of the local that contributes to high performance. Both gradients are essentially complementary to each other and provide a both local and global view of the landscape --- by optimizing $L^{\rm RWP}_{\mathcal{S}}(\boldsymbol{w})$, a good solution can be expected.

Note that, unlike SAM, these two gradient steps are mutually independent, and hence we could efficiently compute them in parallel. In this way, we could ideally halve the training time of SAM. We present the detailed training procedures in \cref{alg:algorithm}.

\begin{algorithm}[ht]
	\caption{RWP algorithm}
	\label{alg:algorithm}
	\KwIn{Loss function $L(\boldsymbol{w})$, initial weight $\boldsymbol{w}_{\rm init}$, training datasets $\mathcal{S}=\{ (\boldsymbol{x}_i,\boldsymbol{y}_i) \}_{i=1}^n$, batch size $b$,  filter number $k$, balance coefficient $\alpha$, noise magnitude $\gamma$, learning rate $\eta$}
	\KwOut{Trained weight $\boldsymbol{w}$}  
	\BlankLine
	Initialize weight $\boldsymbol{w}\gets \boldsymbol{w}_{\rm init}$;\\ \While{\textnormal{not converged}}{
	Sample batch data $\mathcal{B}_1$ and $\mathcal{B}_2$ of size $b$ from $\mathcal{S}$; \\
	Generate random weight perturbations: \\ \quad \quad $\boldsymbol{\epsilon}_r \sim \mathcal{N} \left(0, \gamma \cdot diag \left (\| \boldsymbol{w}_1 \|_2, \cdots ,\| \boldsymbol{w}_k \|_2  \right) \right)$; \\
	Compute the gradients $\boldsymbol{g}_1$ and $\boldsymbol{g}_2$ \emph{in parallel}:\\
	\quad \quad $\boldsymbol{g}_1 \gets \nabla L_{\mathcal{B}_1}(\boldsymbol{w})$, ~$\boldsymbol{g}_2 \gets \nabla L_{\mathcal{B}_2}(\boldsymbol{w} + \bm{\epsilon}_r)$;
	\\
	Update $\boldsymbol{w}$ using gradient descent:\\
	\quad \quad $\boldsymbol{w}\gets \boldsymbol{w}-\eta(\alpha \boldsymbol{g}_1 + (1-\alpha) \boldsymbol{g}_2)$;
	}
	\Return $\boldsymbol{w}$.
\end{algorithm}



\paragraph{Implementation Discussion.}
The parallel computing of two gradients in RWP can be seamlessly integrated into the efficient DDP module in Pytorch \cite{li2020pytorch}.
In \cref{alg:algorithm}, we adopt different batch data for computing $\boldsymbol{g}_1$ and $\boldsymbol{g}_2$ for better adapting to the original DDP pipeline, but it could perform equivalently well using the same batch data. However, for SAM, we find it crucial to use the \emph{same} batch data for the two gradient steps, where different batches will undesirably degenerate the performance to regular SGD training. This justifies the difference of generalization improvement between RWP and SAM: SAM's success closely relies on $m$-sharpness as demonstrated in \cite{andriushchenko2022towards} while RWP is towards a wider flat region guided by random weight perturbations.
Details are elaborated on in Appendix C.

\subsection{Filter Norm Distribution}
We then investigate the filter norm distribution of the models trained by different methods.
First, we review the parameter initialization scheme commonly adopted for DNNs' filters \cite{glorot2010understanding,he2015delving}. For a typical convolution filter $\boldsymbol{w}_k= (C_{\rm in}, H, W)$, the values of initial weights are sampled from a uniform distribution $\mathcal{U}(-\sqrt{t},\sqrt{t})$ with $t=\frac{1}{ C_{\rm in}\times H\times W}$ 
\footnote{\scriptsize Implementation available at
\url{https://pytorch.org/docs/stable/generated/torch.nn.Conv2d.html\#torch.nn.Conv2d}.}. 
Thus, all filters will have a similar initial norm in expectation, since $\mathbb{E} \left[ \| \boldsymbol{w}_k \|^2 \right] = C_{\rm in}\times H\times W \times \frac{1}{12} \left ( 2 \sqrt{t} \right)^2=\frac{1}{3}$.
As the training proceeds, the norms of different filters begin to vary:
some ``winning'' filters that receive larger gradients will evolve faster and result in relatively larger norms, 
while some will be suppressed and gradually shrink under the effect of weight decay. 
Such an in-balance existing in the learning of filters will potentially result in insufficient use of parameters and thus weaken the model's capability.

\begin{figure}[!t]
    \centering
    \includegraphics[width=0.85\linewidth]{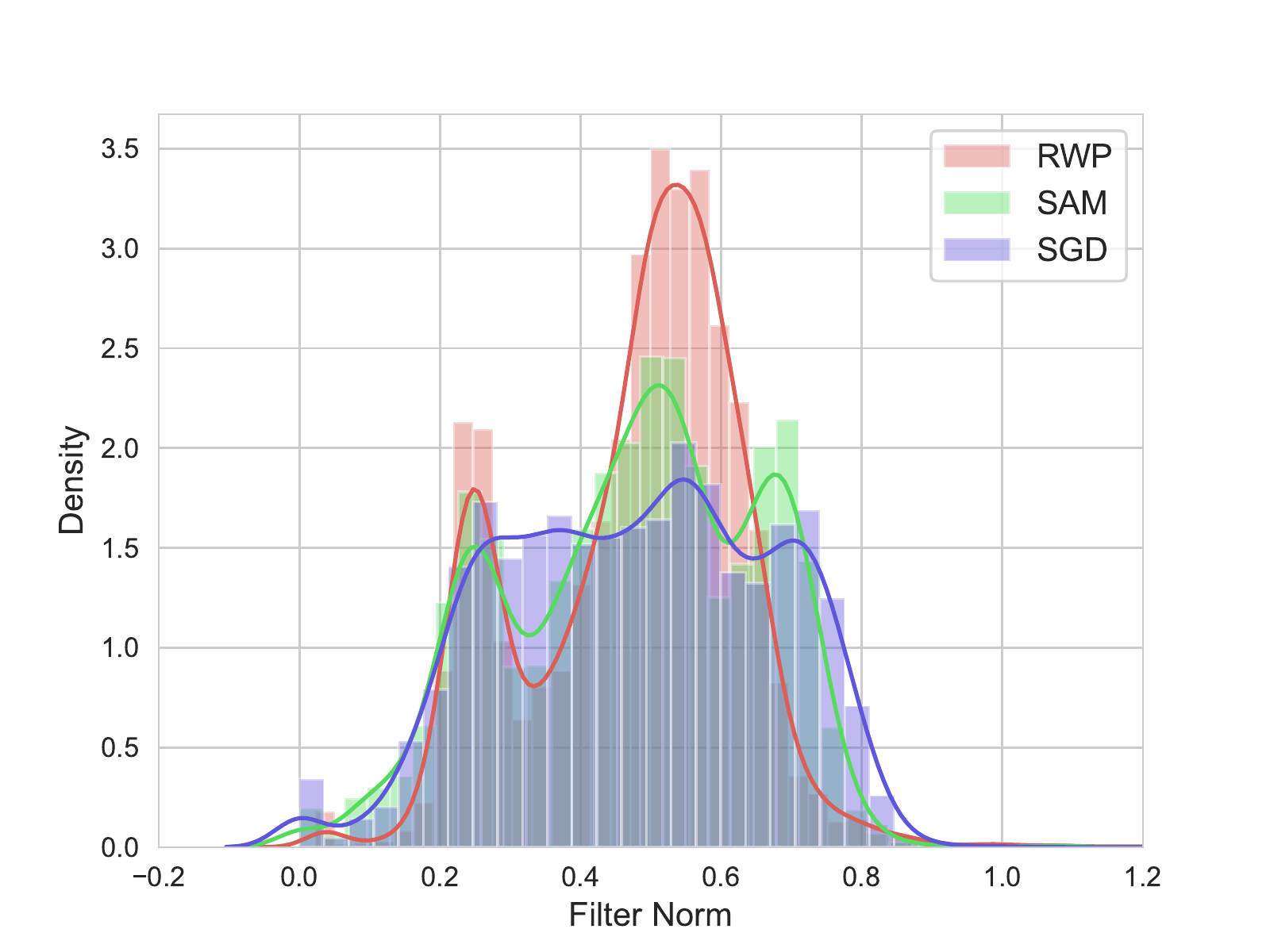}
    \caption{ Distribution of filter norms of ResNet-18 on CIFAR-100. }
    \label{fig:distribution_filternorm}
\end{figure}

 
During the optimization of RWP, the learning of those ``over-growth'' filters will be hindered due to the larger perturbations received, whereas the learning of small-norm filters will be comparatively encouraged with smaller perturbation strength. Such a dynamic regulation will implicitly facilitate balanced learning for different filters and enable more robust representation learning.
In \cref{fig:distribution_filternorm}, we visualize the filter norm distribution of 
models trained by RWP, SAM, and SGD. We observe that the filter norm distribution of RWP is much more concentrated than that of SGD,
indicating more balanced filters learned. This partially explains the success of RWP in generalization improvements, and the improved representation learning will be further verified in \cref{sec:corrupt} with domain-shifted datasets.
We also notice that SAM has similar regularization effects on controlling the filter-norm distribution but RWP's is more obvious.

\section{Experiments}
\label{sec:experiments}

In this section, we conduct extensive experiments to demonstrate the effectiveness of our proposed RWP algorithm. We will first introduce the experimental setup and then evaluate the performance over three benchmark datasets, i.e., CIFAR-10 / 100 and ImageNet. We then show that RWP brings faster convergence and could adapt better to the corruption datasets as well as transfer learning tasks. Finally, we conduct detailed ablation studies to analyze the impact of hyper-parameters.

\subsection{Setup}
\noindent \textbf{Datasets.}
We experiment over three benchmark image classification tasks: CIFAR-10, CIFAR-100 \cite{krizhevsky2009learning} and ImageNet \cite{deng2009imagenet}. For CIFAR, we apply standard preprocessing (including random horizontal flipping, cropping and normalization) and Cutout augmentation \cite{devries2017improved}. For ImageNet, we apply basic data preprocessing and augmentation following the public Pytorch example \cite{paszke2017automatic}.

\noindent \textbf{Models.}
We evaluate a variety of representative architectures, including VGG \cite{simonyan2014very}, ResNet \cite{he2016deep} and Wide-ResNet \cite{ZagoruykoK16}, for both CIFAR and ImageNet datasets. We present additional experiments on ViT \cite{dosovitskiy2020image} in Appendix D.

\noindent \textbf{Training settings.}
We mainly compare the performance of three training schemes: SGD, SAM, and our RWP. 
For CIFAR experiments, we set the training epochs to 200 with batch size 256, momentum 0.9, and weight decay 0.001 \cite{du2022efficient, zhao2022penalizing} for all schemes. 
We set $\rho$ in SAM to 0.05 for CIFAR-10 and 0.10 for CIFAR-100 following previous works \cite{foret2020sharpness,kwon2021asam}.
RWP has two hyper-parameter $\gamma$ and $\alpha$ and we suggest $\gamma=0.01$ and $\alpha=0.5$ for CIFAR. The sensitivity of them could be found in \cref{sec:ablation}.  
We repeat the experiments with 3 independent trials for calculating the mean and standard deviation.
For ImageNet experiments, we set the training epochs to 90 with batch size 256, weight decay 0.0001, and momentum 0.9. We use $\rho=0.05$ for SAM following \cite{foret2020sharpness,kwon2021asam} and $\gamma=0.005$, $\alpha=0.5$ for RWP.
We employ $m$-sharpness with $m=128$ for SAM following \cite{foret2020sharpness,kwon2021asam}. 
For all experiments, we adopt cosine learning rate decay \cite{loshchilov2016sgdr} with an initial learning rate of 0.1 and record the convergence model performance on the test set.
We present a wall-clock time comparison of different methods in Appendix B.

\subsection{CIFAR-10 and CIFAR-100}
\label{sec:cifar}
We first pay attention to the CIFAR-10 and CIFAR-100 datasets. We will consider the final test accuracy reached as well as the total computational cost (FLOPs) and training time for different methods. The detailed comparisons are presented in \cref{tab:cifar}. SAM and RWP involve the same amount of computation, i.e., twice that of SGD, but the training time of RWP could be half that of SAM, due to the good parallelism capability equipped. RWP consistently improves the performance from SGD by $0.6\%$ on CIFAR-10 and $2.6\%$ on CIFAR-100, confirming its effectiveness on generalization improvements.


Besides, we observe that RWP achieves very competitive performance against SAM: for example, it outperforms SAM significantly by $1.0\%$ with VGG16-BN and $0.6\%$ with ResNet-18 on CIFAR-100. On the larger WideResNet models, RWP performs comparably as SAM, e.g., $+0.07\%$ with WideResNet-16-8 and $-0.19\%$ with WideResNet-28-10 on CIFAR-100. 
Note that the parameter size of WideResNet is much larger than that of ResNet-18 and hence training these models is substantially more time-consuming. In this sense, our RWP could be more favorable since it requires only about half of the training time but provides comparable performance.


\begin{table*}[htbp]
 \centering
\caption{Classification accuracy (\%) and training speed comparisons of different methods on the CIFAR-10 and CIFAR-100 datasets.
We set the computation (FLOPs) and training time of regular SGD as the benchmark ($1\times$).
}
 \label{tab:cifar}
 {
\small
 \begin{tabular}{l|c|ccccr}
    \toprule
    Datasets &Model &Training &Epochs &Accuracy &FLOPs &Time \\
    \hline
    \hline
     \multirow{12}{*}{CIFAR-10}  &\multirow{3}{*}{VGG16-BN}  
    &SGD &200   &94.96{\scriptsize $\pm$0.15} &$1\times$ &$\bm 1\times$\\
      &  &RWP  &200   &\textbf{95.61}{\scriptsize $\pm$0.23} &$2\times$ &${\bm  1\times}$\\
    & &SAM &200   &95.43{\scriptsize $\pm$0.11} &$2\times$ &${2\times}$\\
    \cline{2-7}
     &\multirow{3}{*}{ResNet-18} 
     &SGD &200  &96.10{\scriptsize $\pm$0.08} &$1\times$ &$\bm 1\times$\\
     & &RWP &200  &\textbf{96.68}{\scriptsize $\pm$0.17} &$2\times$ &$\bm 1\times$\\   
    & &SAM &200  &96.50{\scriptsize $\pm$0.08} &$2\times$ &${2\times}$\\
    \cline{2-7}
    &\multirow{3}{*}{\shortstack{WRN-16x8}} 
    &SGD &200  &96.81{\scriptsize $\pm$0.10} &$1\times$ &$\bm 1\times$\\
    & & RWP &200  & 97.09{\scriptsize $\pm$0.11} &$2\times$ &$\bm 1\times$\\
    & &SAM &200  & \textbf{97.15}{\scriptsize $\pm$0.10} &$2\times$ 
    &${2\times}$\\
    \cline{2-7}
    &\multirow{3}{*}{WRN-28-10} 
    &SGD &200  &96.85{\scriptsize $\pm$0.05} &$1\times$ &$\bm 1\times$\\
     & & RWP &200  & 97.28{\scriptsize $\pm$0.09} &$2\times$ &$\bm 1\times$\\
    & &SAM &200  & \textbf{97.40}{\scriptsize $\pm$0.06} &$2\times$ & ${2\times}$ \\
    \hline
    \multirow{12}{*}{CIFAR-100}&\multirow{3}{*}{VGG16-BN} 
    &SGD &200   &75.43{\scriptsize $\pm$0.29} &$1\times$ &$\bm 1\times$\\
     & & RWP &200  &\textbf{77.73}{\scriptsize$\pm$0.05} &$2\times$ &$\bm 1\times$\\
    & &SAM &200  &76.74{\scriptsize $\pm$0.22} &$2\times$ &${2\times}$\\
    \cline{2-7}
    &\multirow{3}{*}{ResNet-18} 
    &SGD &200   &78.10{\scriptsize $\pm$0.39} &$1\times$ &$\bm 1\times$\\
    & & RWP &200  &\textbf{80.75}{\scriptsize$\pm$0.16} &$2\times$ &$\bm 1\times$\\
    & &SAM &200   &80.16{\scriptsize$\pm$0.25} &$2\times$ &${2\times}$\\
    \cline{2-7}
    &\multirow{3}{*}{WRN-16-8} 
    &SGD &200 &81.39{\scriptsize$\pm$0.10} &$1\times$ &$\bm 1\times$\\
    & & RWP &200  & \textbf{83.52}{\scriptsize$\pm$0.20} &$2\times$ &$ \bm 1\times$\\
    & &SAM &200  & 83.45{\scriptsize$\pm$0.26} &$2\times$ &${2\times}$\\
    \cline{2-7}
    &\multirow{3}{*}{WRN-28-10} 
    &SGD &200 & 82.51{\scriptsize$\pm$0.24} &$1\times$ &$\bm 1\times$\\
    & & RWP &200  & 84.25{\scriptsize$\pm$0.12} &$2\times$ &$\bm 1\times$\\
    & &SAM &200  & \textbf{84.44}{\scriptsize$\pm$0.03} &$2\times$ & ${2\times}$ \\
    \bottomrule
 \end{tabular}
 }
\end{table*}

\subsection{ImageNet}
\label{sec:imagenet}
We then move on to the ImageNet, of which the scale is substantially larger. Thus, $2\times$ training time required in SAM would be prohibitively slow, and improving the training efficiency is of great significance.  
Here, we apply SGD, SAM and RWP to three different architectures, VGG16-BN, ResNet-18 and ResNet-50. The results are illustrated in \cref{tab:ImageNet}. 
We observe that RWP can outperform the SAM by a large margin:
without any other enhancing technique such as label smoothing \cite{szegedy2016rethinking} or CutMix \cite{yun2019cutmix}, we achieve $75.79\%$ accuracy with VGG16-BN, $71.58\%$ accuracy with ResNet-18 and $78.04\%$ accuracy with ResNet-50, surpassing SAM as large as $1.1\%$. Importantly, we achieve new state-of-the-art results of SAM's follow-up works on the ImageNet, and these results further demonstrate the effectiveness of RWP over large-scale problems.

\subsection{Faster Convergence}
We visualize the training loss and test accuracy curves of SGD, SAM, and RWP on CIFAR-100 in \cref{fig:training}. Training curves of other datasets can be found in Appendix A. We observe that RWP could achieve a much faster convergence than SAM and SGD, which is shown in both training loss and test accuracy. This demonstrates that the active cooperation of the original gradient and the perturbed gradient in RWP could benefit optimization as well as generalization. 

\begin{figure}[!b]
 \centering
 \begin{subfigure}{0.495\linewidth}
 \centering
  \includegraphics[width=1\linewidth]{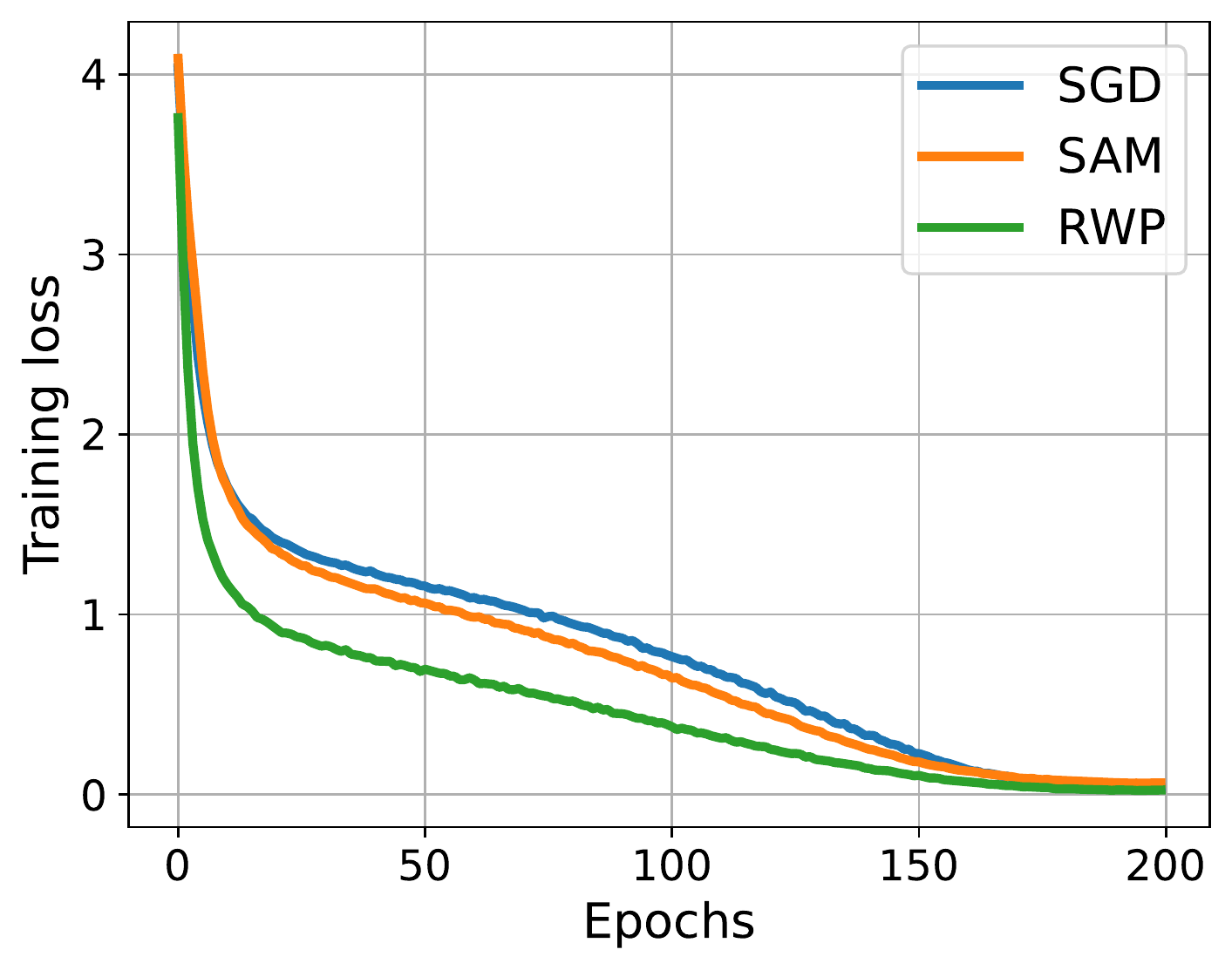}
    \caption{Training loss}
 \end{subfigure}
 \begin{subfigure}{0.495\linewidth}
 \centering
  \includegraphics[width=1\linewidth]{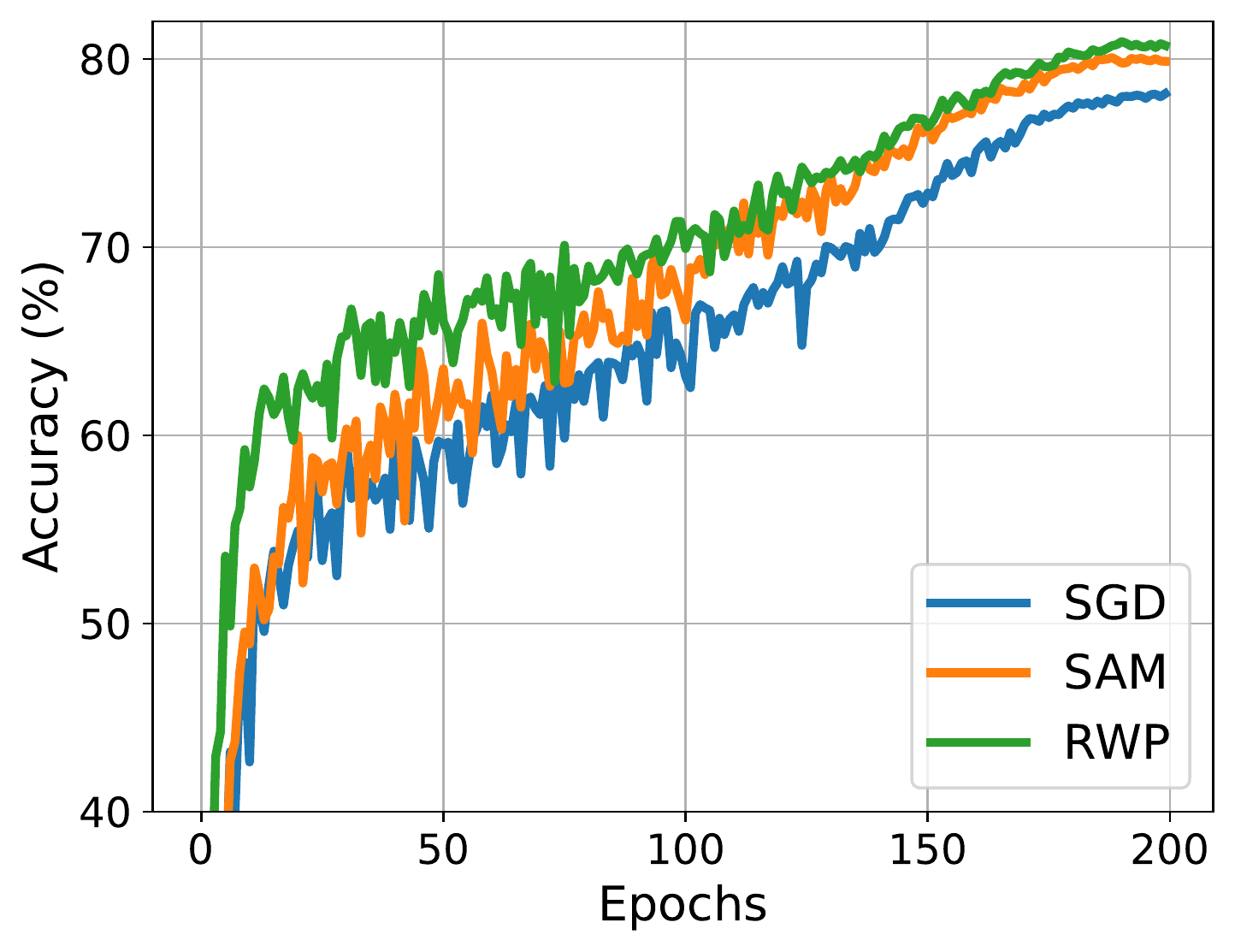}
    \caption{Test accuracy}
 \end{subfigure}
 \caption{Training curves on CIFAR-100 with ResNet-18 model.}
 \label{fig:training}
\end{figure}

As RWP could achieve a faster convergence, we expect that it would attain more benefits when the training epochs is limited.
To demonstrate this, we further reduce the training epochs of SAM and RWP to 100 epochs, while keeping 200 epochs for SGD. In this scenario, all methods take the same amount of computation but RWP achieves the \emph{fastest} training speed (i.e., $\mathbf{ 0.5\times}$ training time
as SGD and SAM
). Similarly, we report the performance comparisons of different methods in \cref{tab:cifar100} and observe that RWP consistently outperforms SAM and SGD over 
all four architectures. Especially on VGG16-BN, SAM performs slightly worse than SGD
perhaps due to the fewer iterations taken, 
but RWP still surpasses SGD by a large margin (i.e. $0.8\%$) and attains over $76\%$ test accuracy.
This further confirms the fast convergence of RWP, demonstrating that it could achieve effective generalization improvements efficiently.

\begin{table}[!h]
    \centering
    \small
    \caption{Performance comparisons on CIFAR-100. We train 100 epochs for SAM and RWP and 200 epochs for SGD. All methods consumed the \textbf{same} computation and RWP manifests the best performance as well as training speed.}
    \label{tab:cifar100}
    \begin{tabular}{l|cccr}
    \toprule
         Model &Training &Accuracy &FLOPs &Time\\
    \hline
    \hline
         \multirow{3}{*}{VGG16-BN} 
         & RWP &\textbf{76.21}{\scriptsize$\pm$0.05} &$1\times$ &$\bm{0.5\times}$\\
         &SGD &75.43{\scriptsize$\pm$0.28} &$1\times$ &$1\times$\\
         &SAM &75.24{\scriptsize$\pm$0.28} &$1\times$ &$1\times$\\
    \cline{1-5}
         \multirow{3}{*}{ResNet-18} 
         & RWP &\textbf{79.95}{\scriptsize$\pm$0.20} &$1\times$ &$\bm{0.5\times}$\\
         &SGD &78.10{\scriptsize$\pm$0.39} &$1\times$ &$1\times$\\
         &SAM &79.42{\scriptsize$\pm$0.33} &$1\times$ &$1\times$\\
    \cline{1-5}
         \multirow{3}{*}{WRN-16-8} 
         & RWP &\textbf{82.64}{\scriptsize$\pm$0.23} &$1\times$ &$\bm{0.5\times}$\\
         &SGD &81.39{\scriptsize$\pm$0.10} &$1\times$ &$1\times$\\
         &SAM &82.49{\scriptsize$\pm$0.27} &$1\times$ &$1\times$\\
    \cline{1-5}
         \multirow{3}{*}{WRN-28-10} 
         & RWP 
         &\textbf{83.58}{\scriptsize$\pm$0.13} &$1\times$ &$\bm {0.5\times}$\\
         &SGD &82.51{\scriptsize$\pm$0.24} &$1\times$ &$1\times$\\
         &SAM &83.28{\scriptsize$\pm$0.12} &$1\times$ &$1\times$\\
    \bottomrule
    \end{tabular}
\end{table}


\subsection{Robustness to Common Corruption}
\label{sec:corrupt}
\begin{figure}[!t]
 \centering
 \begin{subfigure}{0.495\linewidth}
 \centering
  \includegraphics[width=1\linewidth]{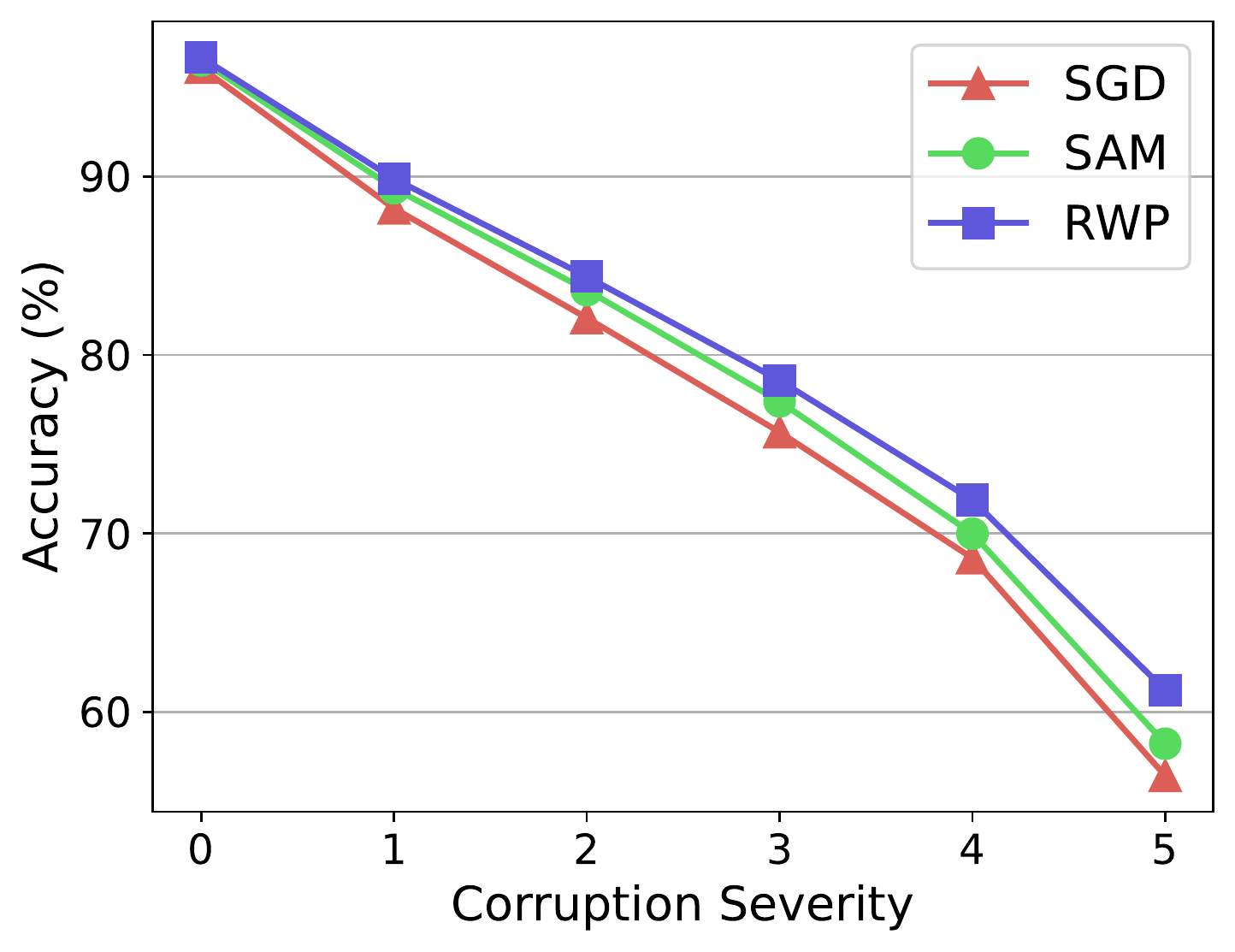}
    \caption{ResNet-18 on CIFAR-10-C.}
 \end{subfigure}
 \begin{subfigure}{0.495\linewidth}
 \centering
  \includegraphics[width=1\linewidth]{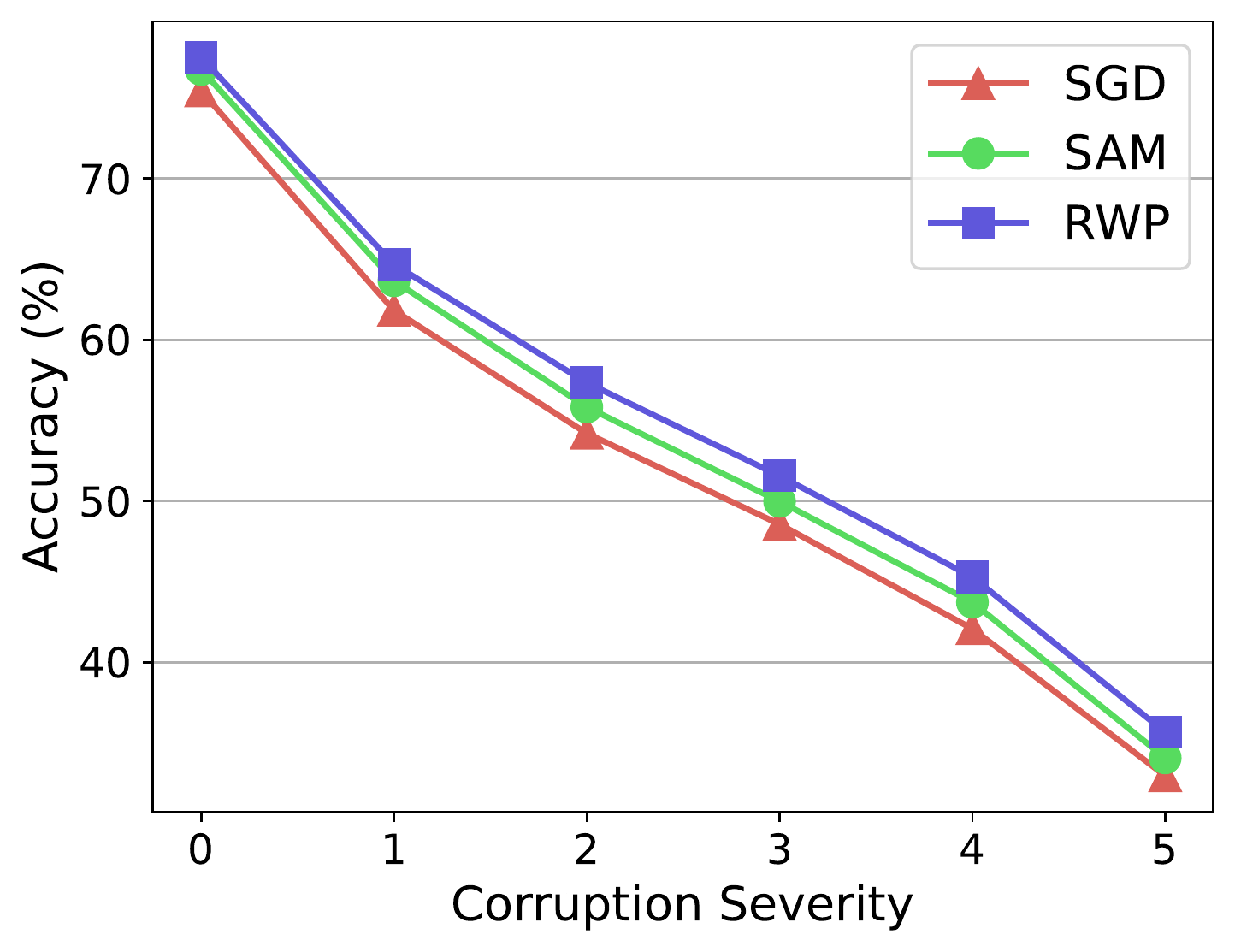}
    \caption{VGG16-BN on CIFAR-100-C.}
 \end{subfigure}
 \caption{Test accuracy curves w.r.t. different levels of corruptions on CIFAR-10-C and CIFAR-100-C.} \label{fig:corruptions}
\end{figure}

Compared with SAM, RWP imposes a significantly larger magnitude of weight perturbations and attains a landscape with a wider flat region (as demonstrated in \cref{fig:visualization}). Hence, we expect that it could adapt better to the common corruption on data, especially severe perturbations. 
To this end, we evaluate over CIFAR-10-C and CIFAR-100-C datasets \cite{hendrycks2019robustness}, which are constructed by corrupting the original CIFAR-10 / 100 test sets. There are in total 15 kinds of noise categorized into three types, i.e., blur, weather, and digital corruption. Each kind of corruption has 5 levels of severity.
Here, we first test the performance of the models obtained from \cref{sec:cifar} at the most severe level of corruption (i.e. level 5). We calculate the averaged test accuracies over all 15 corruptions and report them in \cref{tab:cifar-c}. 
We observe that RWP consistently outperforms the SAM counterparts by $3.0\%$ on CIFAR-10-C and $2.3\%$ on CIFAR-100-C, of which the advantages are much more obvious than that on the clean test set. Note that during the training, these corrupted data are not touched and hence the robustness to data corruptions is naturally benefited from the wider flat minimum that RWP's optimization brings.
Then in \cref{fig:corruptions}, we investigate the performance under different severity levels of corruption. We adopt ResNet-18 for CIFAR-10-C and VGG16-BN for CIFAR-100-C and plot the test accuracy curves w.r.t. different severity levels. We observe that RWP consistently outperforms SGD and SAM, and such advantages are more obvious with higher corruption severity.


\begin{table}[htbp]
    \centering
    \small
    \caption{Classification accuracy (\%) comparisons on CIFAR-10-C and CIFAR-100-C datasets with different methods at the most severe level of corruption (level 5).   }
    \label{tab:cifar-c}
    \begin{tabular}{lc|ccc}
    \toprule
         Datasets &Model &SGD &SAM &RWP \\
    \hline
    \hline
         \multirow{4}{*}{CIFAR-10-C} &VGG16-BN &56.71 &57.82 &\textbf{60.23} \\
         &ResNet-18 &56.42 &58.22 &\textbf{61.21} \\
         &WRN-16-8 &57.90 &60.09 &\textbf{60.56} \\
         &WRN-28-10 &58.63 &61.84 &\textbf{62.13}
\\
    \hline
         \multirow{4}{*}{CIFAR-100-C} &VGG16-BN &32.96 &34.07 &\textbf{35.66}\\
         &ResNet-18 &31.26 &32.64 &\textbf{34.97}\\
         &WRN-16-8 &33.51 &35.39 &\textbf{36.27}\\
         &WRN-28-10 &35.49 &37.29 &\textbf{37.71}\\
    
    \bottomrule
    \end{tabular}
\end{table}

\subsection{Transfer Learning by Fine-tuning}
We apply RWP to fine-tuning tasks, where we use the vision transformer model DeiT-small\cite{deit} pre-trained on the ImageNet and finetune it on the CIFAR-10 / 100. We use Adam \cite{kingma2014adam} as the base optimizer and finetune the models for 20 epochs with batch size 128 and an initial learning rate of 0.0001. Detailed settings could be found in Appendix E. From the results in \cref{tab:fintune}, we observe RWP consistently improves the performance over the competing methods.

\begin{table}[htbp]
    \centering
    \small
    \begin{tabular}{l|ccc}
    \toprule
     Datasets   & Adam  & Adam-SAM   & Adam-RWP \\ \hline \hline
     CIFAR-10   &   98.08{\scriptsize$\pm$0.05}    & 97.99{\scriptsize$\pm$0.10} & \textbf{98.21}{\scriptsize$\pm$0.10}   \\  
     CIFAR-100  & 88.64{\scriptsize$\pm$0.14} & 88.74{\scriptsize$\pm$0.13} & \textbf{89.01}{\scriptsize$\pm$0.21} \\
    \bottomrule
    \end{tabular}
    \caption{Test accuracies for fine-tuning}
    \label{tab:fintune}
\end{table}

\begin{table*}[!t]
 \centering
 \caption{
 Classification accuracies (\%) and training speed comparisons of different methods on the ImageNet datasets.
We set the computation (FLOPs) and training time of regular SGD as the benchmark ($1\times$).
 } 
 \label{tab:ImageNet}
 {
\small
 \begin{tabular}{l|cccccr}
    \toprule
    Model &Training &Epochs &Top-1 Accuracy &Top-5 Accuracy &FLOPs &Time \\
    \hline
    \hline
    \multirow{3}{*}{VGG16-BN} &SGD &90 &73.11 &91.12  &$1\times$ &$\bm 1\times$\\
    &  RWP &90 &\textbf{75.79} &\textbf{92.92} &$2\times$ &$\bm 1\times$\\    
    &SAM &90 &74.65 &92.21 &$2\times$ &${2\times}$\\
        \hline
    \multirow{3}{*}{ResNet-18} &SGD &90 &70.46 &89.79 &$2\times$ &$\bm 1\times$\\
    & RWP &90 &\textbf{71.58} &\textbf{90.31} &$2\times$ &$\bm 1\times$\\
    &SAM &90 &70.77 &89.83 &$2\times$ &${2\times}$\\
        \hline
    \multirow{3}{*}{ResNet-50} &SGD &90 &76.83 &93.55  &$2\times$ &$\bm 1\times$\\
    & RWP &90 &\textbf{78.04} &\textbf{93.91} &$2\times$ &$\bm 1\times$\\
    &SAM &90 &77.15 &{93.55} &$2\times$ &${2\times}$\\    
        \bottomrule
 \end{tabular}
 }
\end{table*}




\subsection{Ablation Studies}
\label{sec:ablation}
In our RWP, there are two hyperparameters: the balance coefficients $\alpha$ and the perturbation magnitude $\gamma$. To better understand the impact of these hyperparameters on model performance, we conduct two sets of ablation studies with two representative architectures, i.e., VGG16-BN and ResNet-18, on the CIFAR-100 datasets.

\begin{figure}[htbp]
    \centering
    \includegraphics[width=0.8\linewidth]{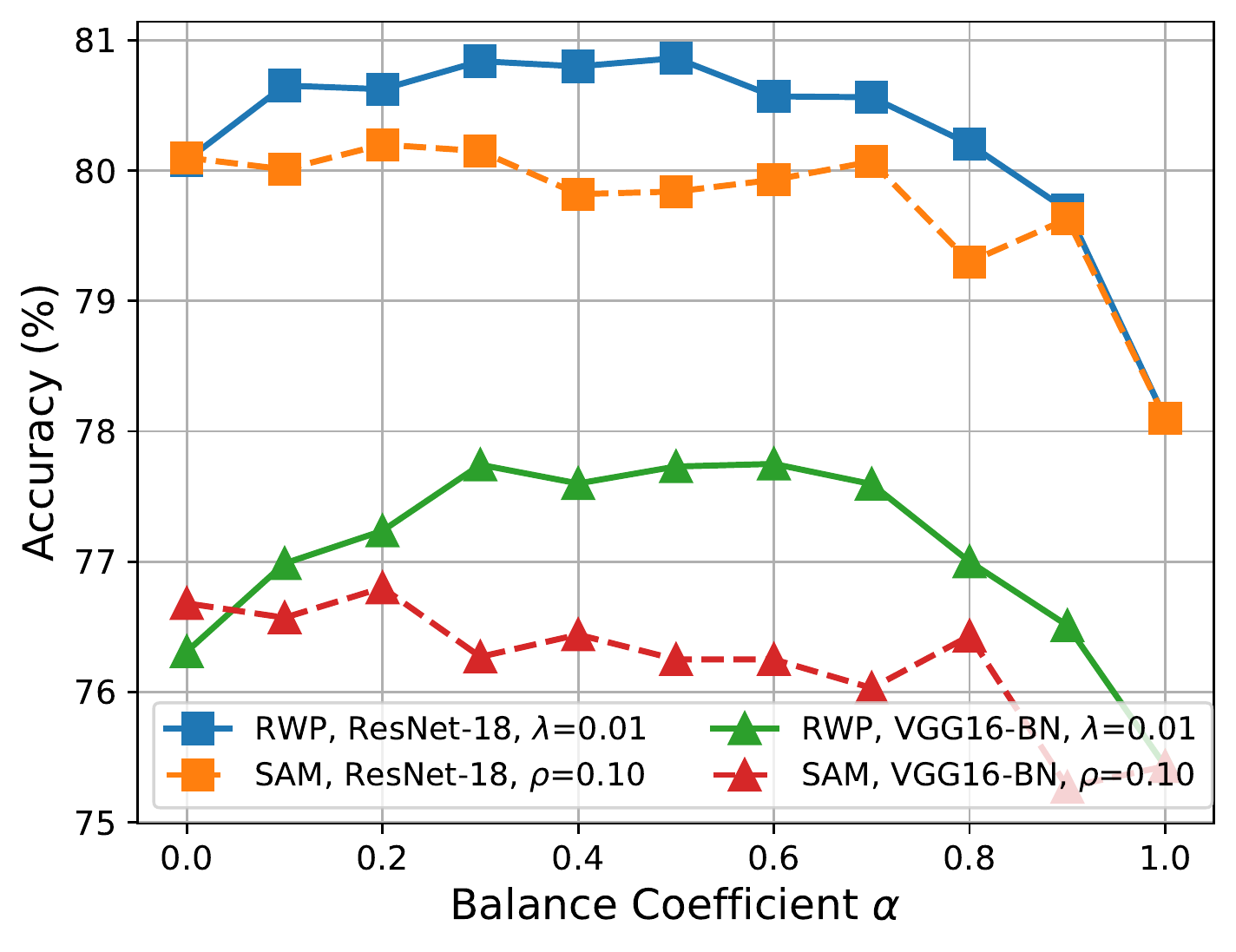}
    \caption{Hyperparameter sensitivity of $\alpha$.}
    \label{fig:comparison_alpha}
\end{figure}

We first focus on the balance coefficient $\alpha$. To investigate its impact, we 
traverse over the interval $[0,1]$ with a step size of $0.1$. The results are plotted in \cref{fig:comparison_alpha}. As we could see in the figure, from the standard SGD training scheme $(\alpha=1)$ to the pure random perturbation scheme $(\alpha=0)$, the test accuracy curves for RWP can experience a first increase and then a decrease, where the optimal is obtained roughly at $\alpha=0.5$. Clearly, in our cases, minimizing either the perturbed loss function $(\alpha=0)$ or the original loss function $(\alpha=1)$ alone can not produce the best performance. By jointly optimizing them, we could take advantage of both and thereby significantly enhance the performance. A minor finding is that $\alpha$ is not that sensitive and could yield similar performance in the range of $[0.3, 0.8]$.

\noindent
\textbf{SAM with mixing gradients}.
Similar to RWP, SAM also calculates two gradients for each update step: $\nabla L(\boldsymbol{w})$ and $\nabla L(\boldsymbol{w} + \bm{\epsilon}_s)$. It is natural to wonder whether mixing these two gradient steps would bring similar accuracy improvements since it virtually involves no additional computation burden. In \cref{fig:comparison_alpha}, we also plot the performance of SAM updated with mixing gradients: $\alpha \nabla L(\boldsymbol{w}) + (1 - \alpha) L(\boldsymbol{w} + \bm{\epsilon}_s)$, where $\alpha$ is the balance coefficient the same as in RWP. We observe that mixing gradients indeed brings accuracy improvements (best at nearly $\alpha=0.2$), but it is very minor (e.g. $+0.12\%$) compared with the improvements in RWP. This is perhaps due to the perturbation magnitude in SAM being relatively small, which wouldn't severely ``blur'' the loss function, and introducing $\nabla L(\boldsymbol{w})$ would conversely degrade the regularization effect.

\begin{figure}[htbp]
    \centering
    \includegraphics[width=0.8\linewidth]{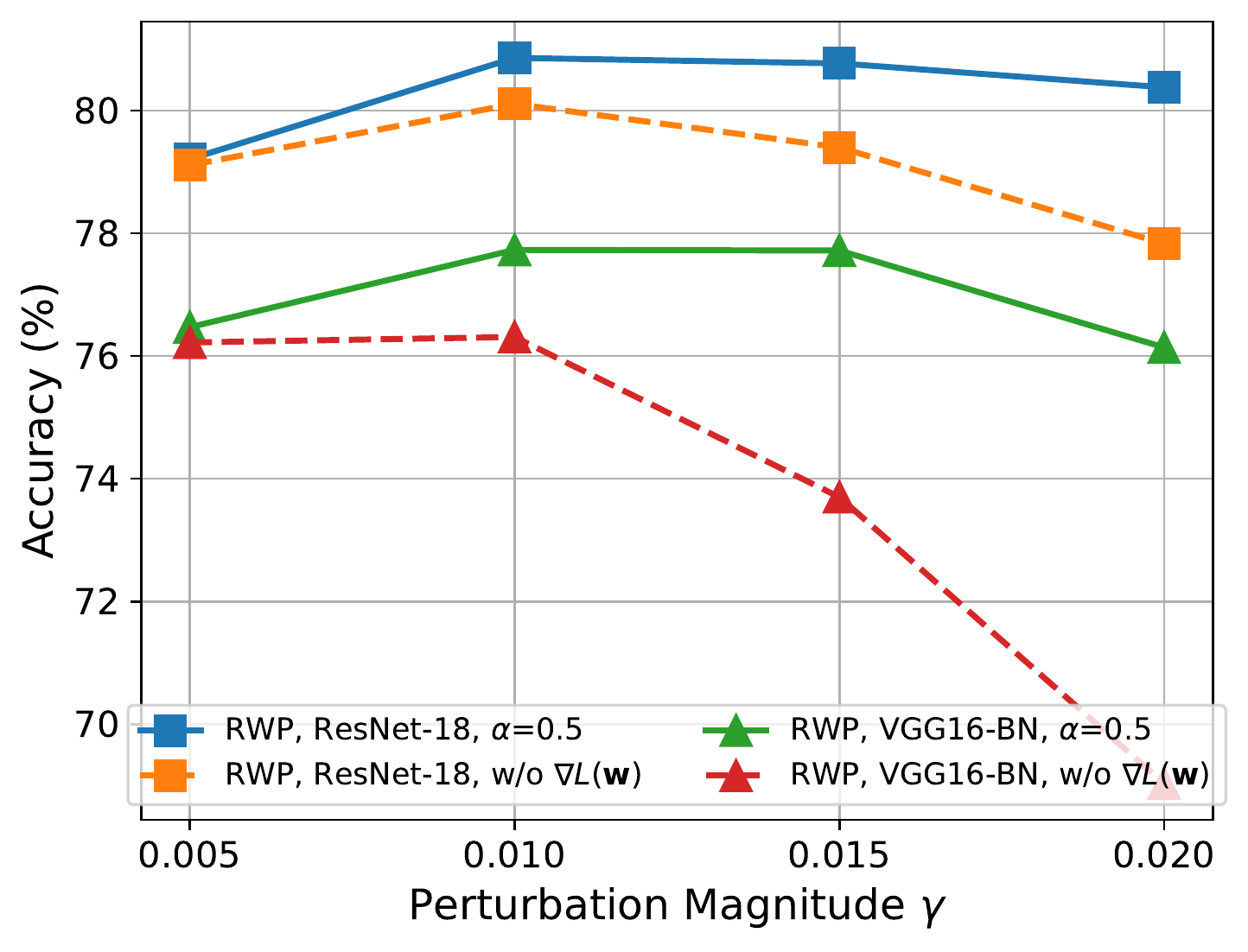}
    \caption{Hyperparameter sensitivity of $\gamma$.}
    \label{fig:comparison_rho}
\end{figure}

We then study the impact of perturbation magnitude $\gamma$. We keep the balance coefficient $\alpha$ to 0.5 as determined in the above and vary $\gamma$ in the set $\{0.005, 0.01, 0.015, 0.02\}$. In \cref{fig:comparison_rho}, we observe that optimal magnitude is reached with $\gamma=0.01$ for both architectures, while too large/small perturbation strength would degrade the performance. 

\noindent
\textbf{RWP without $\nabla L(\boldsymbol{w})$}.
In the above, we have shown that the perturbed gradient $\nabla L(\boldsymbol{w}+\bm{\epsilon}_r)$ alone can not yield the best performance. We then vary its perturbation magnitude $\gamma$ and plot the accuracy curves with dotted line in \cref{fig:comparison_rho}. We obverse that varying $\gamma$ does not bring further improvements, whereas the fluctuation of performance w.r.t. $\gamma$ would be much more severe compared with RWP with $\nabla L(\boldsymbol{w})$.

\section{Conclusion and Future Works}
\label{sec:conclusion}

In this work, we introduce RWP, a simple training algorithm that jointly optimizes the original loss function and the loss function with random weight perturbations, to achieve efficient generalization improvement. Different from SAM, the two gradient steps involved in RWP are separable and thus could be efficiently computed in parallel,  requiring only about half of the training time of SAM.
We conduct extensive experiments with various architectures and datasets to demonstrate the superior performance of RWP, where several new state-of-the-art performances are achieved. We also show that RWP exhibits faster convergence and adapts better to data corruption and fine-tuning tasks. Future works include designing more effective schemes for random perturbation generation (e.g. considering the structure/layer information) and applying RWP to other training schemes (e.g. large-batch training). 

{\small
\bibliographystyle{ieee_fullname}
\bibliography{main}

\begin{thebibliography}{10}\itemsep=-1pt

\bibitem{andriushchenko2022towards}
Maksym Andriushchenko and Nicolas Flammarion.
\newblock Towards understanding sharpness-aware minimization.
\newblock In {\em International Conference on Machine Learning (ICML)}, 2022.

\bibitem{ba2016layer}
Jimmy~Lei Ba, Jamie~Ryan Kiros, and Geoffrey~E Hinton.
\newblock Layer normalization.
\newblock {\em arXiv preprint arXiv:1607.06450}, 2016.

\bibitem{bisla2022low}
Devansh Bisla, Jing Wang, and Anna Choromanska.
\newblock Low-pass filtering sgd for recovering flat optima in the deep
  learning optimization landscape.
\newblock In {\em International Conference on Artificial Intelligence and
  Statistics (AISTATIS)}, 2022.

\bibitem{chaudhari2017entropy}
Pratik Chaudhari, Anna Choromanska, Stefano Soatto, Yann LeCun, Carlo Baldassi,
  Christian Borgs, Jennifer Chayes, Levent Sagun, and Riccardo Zecchina.
\newblock Entropy-sgd: Biasing gradient descent into wide valleys.
\newblock In {\em International Conference on Learning Representations (ICLR)},
  2017.

\bibitem{cubuk2019autoaugment}
Ekin~D Cubuk, Barret Zoph, Dandelion Mane, Vijay Vasudevan, and Quoc~V Le.
\newblock Autoaugment: Learning augmentation strategies from data.
\newblock In {\em Proceedings of the IEEE Conference on Computer Vision and
  Pattern Recognition (CVPR)}, 2019.

\bibitem{cubuk2020randaugment}
Ekin~D Cubuk, Barret Zoph, Jonathon Shlens, and Quoc~V Le.
\newblock Randaugment: Practical automated data augmentation with a reduced
  search space.
\newblock In {\em Proceedings of the IEEE/CVF Conference on Computer Vision and
  Pattern Recognition Workshops}, pages 702--703, 2020.

\bibitem{deng2009imagenet}
Jia Deng, Wei Dong, Richard Socher, Li-Jia Li, Kai Li, and Li Fei-Fei.
\newblock Imagenet: A large-scale hierarchical image database.
\newblock In {\em Proceedings of the IEEE Conference on Computer Vision and
  Pattern Recognition (CVPR)}, 2009.

\bibitem{devries2017improved}
Terrance DeVries and Graham~W Taylor.
\newblock Improved regularization of convolutional neural networks with cutout.
\newblock {\em arXiv preprint arXiv:1708.04552}, 2017.

\bibitem{dinh2017sharp}
Laurent Dinh, Razvan Pascanu, Samy Bengio, and Yoshua Bengio.
\newblock Sharp minima can generalize for deep nets.
\newblock In {\em International Conference on Machine Learning (ICML)}, 2017.

\bibitem{dosovitskiy2020image}
Alexey Dosovitskiy, Lucas Beyer, Alexander Kolesnikov, Dirk Weissenborn,
  Xiaohua Zhai, Thomas Unterthiner, Mostafa Dehghani, Matthias Minderer, Georg
  Heigold, Sylvain Gelly, et~al.
\newblock An image is worth 16x16 words: Transformers for image recognition at
  scale.
\newblock In {\em International Conference on Learning Representations (ICLR)},
  2021.

\bibitem{du2022efficient}
Jiawei Du, Hanshu Yan, Jiashi Feng, Joey~Tianyi Zhou, Liangli Zhen, Rick
  Siow~Mong Goh, and Vincent~YF Tan.
\newblock Efficient sharpness-aware minimization for improved training of
  neural networks.
\newblock In {\em International Conference on Learning Representations (ICLR)},
  2022.

\bibitem{du2022sharpness}
Jiawei Du, Daquan Zhou, Jiashi Feng, Vincent~YF Tan, and Joey~Tianyi Zhou.
\newblock Sharpness-aware training for free.
\newblock {\em arXiv preprint arXiv:2205.14083}, 2022.

\bibitem{foret2020sharpness}
Pierre Foret, Ariel Kleiner, Hossein Mobahi, and Behnam Neyshabur.
\newblock Sharpness-aware minimization for efficiently improving
  generalization.
\newblock In {\em International Conference on Learning Representations (ICLR)},
  2020.

\bibitem{glorot2010understanding}
Xavier Glorot and Yoshua Bengio.
\newblock Understanding the difficulty of training deep feedforward neural
  networks.
\newblock In {\em International Conference on Artificial Intelligence and
  Statistics (AISTATIS)}, 2010.

\bibitem{he2015delving}
Kaiming He, Xiangyu Zhang, Shaoqing Ren, and Jian Sun.
\newblock Delving deep into rectifiers: Surpassing human-level performance on
  imagenet classification.
\newblock In {\em Proceedings of the IEEE Conference on Computer Vision and
  Pattern Recognition (CVPR)}, 2015.

\bibitem{he2016deep}
Kaiming He, Xiangyu Zhang, Shaoqing Ren, and Jian Sun.
\newblock Deep residual learning for image recognition.
\newblock In {\em Proceedings of the IEEE Conference on Computer Vision and
  Pattern Recognition (CVPR)}, 2016.

\bibitem{hendrycks2019robustness}
Dan Hendrycks and Thomas Dietterich.
\newblock Benchmarking neural network robustness to common corruptions and
  perturbations.
\newblock In {\em International Conference on Learning Representations (ICLR)},
  2019.

\bibitem{hochreiter1997flat}
Sepp Hochreiter and J{\"u}rgen Schmidhuber.
\newblock Flat minima.
\newblock {\em Neural computation}, 1997.

\bibitem{ioffe2015batch}
Sergey Ioffe and Christian Szegedy.
\newblock Batch normalization: Accelerating deep network training by reducing
  internal covariate shift.
\newblock In {\em International Conference on Machine Learning (ICML)}, 2015.

\bibitem{ishida2020we}
Takashi Ishida, Ikko Yamane, Tomoya Sakai, Gang Niu, and Masashi Sugiyama.
\newblock Do we need zero training loss after achieving zero training error?
\newblock In {\em International Conference on Machine Learning (ICML)}, 2020.

\bibitem{izmailov2018averaging}
Pavel Izmailov, Dmitrii Podoprikhin, Timur Garipov, Dmitry Vetrov, and
  Andrew~Gordon Wilson.
\newblock Averaging weights leads to wider optima and better generalization.
\newblock {\em arXiv preprint arXiv:1803.05407}, 2018.

\bibitem{keskar2017large}
Nitish~Shirish Keskar, Dheevatsa Mudigere, Jorge Nocedal, Mikhail Smelyanskiy,
  and Ping Tak~Peter Tang.
\newblock On large-batch training for deep learning: Generalization gap and
  sharp minima.
\newblock In {\em International Conference on Learning Representations (ICLR)},
  2017.

\bibitem{kim2022fisher}
Minyoung Kim, Da Li, Shell~X Hu, and Timothy Hospedales.
\newblock Fisher sam: Information geometry and sharpness aware minimisation.
\newblock In {\em International Conference on Machine Learning (ICML)}, 2022.

\bibitem{kingma2014adam}
Diederik~P. {Kingma} and Jimmy~Lei {Ba}.
\newblock Adam: A method for stochastic optimization.
\newblock In {\em International Conference on Learning Representations (ICLR)},
  2015.

\bibitem{kolesnikov2020big}
Alexander Kolesnikov, Lucas Beyer, Xiaohua Zhai, Joan Puigcerver, Jessica Yung,
  Sylvain Gelly, and Neil Houlsby.
\newblock Big transfer (bit): General visual representation learning.
\newblock In {\em European conference on computer vision (ECCV)}, 2020.

\bibitem{krizhevsky2009learning}
Alex Krizhevsky and Geoffrey Hinton.
\newblock Learning multiple layers of features from tiny images.
\newblock {\em Technical Report}, 2009.

\bibitem{krogh1991simple}
Anders Krogh and John Hertz.
\newblock A simple weight decay can improve generalization.
\newblock In {\em Advances in Neural Information Processing Systems (NeurIPS)},
  1991.

\bibitem{kwon2021asam}
Jungmin Kwon, Jeongseop Kim, Hyunseo Park, and In~Kwon Choi.
\newblock Asam: Adaptive sharpness-aware minimization for scale-invariant
  learning of deep neural networks.
\newblock In {\em International Conference on Machine Learning (ICML)}, 2021.

\bibitem{li2018visualizing}
Hao Li, Zheng Xu, Gavin Taylor, Christoph Studer, and Tom Goldstein.
\newblock Visualizing the loss landscape of neural nets.
\newblock In {\em Advances in Neural Information Processing Systems (NeurIPS)},
  2018.

\bibitem{li2020pytorch}
Shen Li, Yanli Zhao, Rohan Varma, Omkar Salpekar, Pieter Noordhuis, Teng Li,
  Adam Paszke, Jeff Smith, Brian Vaughan, Pritam Damania, et~al.
\newblock Pytorch distributed: Experiences on accelerating data parallel
  training.
\newblock {\em arXiv preprint arXiv:2006.15704}, 2020.

\bibitem{liu2022towards}
Yong Liu, Siqi Mai, Xiangning Chen, Cho-Jui Hsieh, and Yang You.
\newblock Towards efficient and scalable sharpness-aware minimization.
\newblock In {\em Proceedings of the IEEE/CVF Conference on Computer Vision and
  Pattern Recognition}, 2022.

\bibitem{liu2021swin}
Ze Liu, Yutong Lin, Yue Cao, Han Hu, Yixuan Wei, Zheng Zhang, Stephen Lin, and
  Baining Guo.
\newblock Swin transformer: Hierarchical vision transformer using shifted
  windows.
\newblock In {\em Proceedings of the IEEE/CVF International Conference on
  Computer Vision (CVPR)}, 2021.

\bibitem{loshchilov2016sgdr}
Ilya Loshchilov and Frank Hutter.
\newblock Sgdr: Stochastic gradient descent with warm restarts.
\newblock {\em arXiv preprint arXiv:1608.03983}, 2016.

\bibitem{mi2022make}
Peng Mi, Li Shen, Tianhe Ren, Yiyi Zhou, Xiaoshuai Sun, Rongrong Ji, and
  Dacheng Tao.
\newblock Make sharpness-aware minimization stronger: A sparsified perturbation
  approach.
\newblock {\em arXiv preprint arXiv:2210.05177}, 2022.

\bibitem{neyshabur2017exploring}
Behnam Neyshabur, Srinadh Bhojanapalli, David McAllester, and Nati Srebro.
\newblock Exploring generalization in deep learning.
\newblock {\em Advances in neural information processing systems (NeurIPS)},
  2017.

\bibitem{paszke2017automatic}
Adam Paszke, Sam Gross, Soumith Chintala, Gregory Chanan, Edward Yang, Zachary
  DeVito, Zeming Lin, Alban Desmaison, Luca Antiga, and Adam Lerer.
\newblock Automatic differentiation in pytorch.
\newblock 2017.

\bibitem{radford2021learning}
Alec Radford, Jong~Wook Kim, Chris Hallacy, Aditya Ramesh, Gabriel Goh,
  Sandhini Agarwal, Girish Sastry, Amanda Askell, Pamela Mishkin, Jack Clark,
  et~al.
\newblock Learning transferable visual models from natural language
  supervision.
\newblock In {\em International Conference on Machine Learning (ICML)}, 2021.

\bibitem{simonyan2014very}
Karen Simonyan and Andrew Zisserman.
\newblock Very deep convolutional networks for large-scale image recognition.
\newblock {\em arXiv preprint arXiv:1409.1556}, 2014.

\bibitem{srivastava2014dropout}
Nitish Srivastava, Geoffrey Hinton, Alex Krizhevsky, Ilya Sutskever, and Ruslan
  Salakhutdinov.
\newblock Dropout: a simple way to prevent neural networks from overfitting.
\newblock {\em Journal of Machine Learning Research (JMLR)}, 15(1):1929--1958,
  2014.

\bibitem{szegedy2016rethinking}
Christian Szegedy, Vincent Vanhoucke, Sergey Ioffe, Jon Shlens, and Zbigniew
  Wojna.
\newblock Rethinking the inception architecture for computer vision.
\newblock In {\em Proceedings of the IEEE Conference on Computer Vision and
  Pattern Recognition (CVPR)}, 2016.

\bibitem{tan2019efficientnet}
Mingxing Tan and Quoc Le.
\newblock Efficientnet: Rethinking model scaling for convolutional neural
  networks.
\newblock In {\em International Conference on Machine Learning (ICML)}, 2019.

\bibitem{deit}
Hugo Touvron, Matthieu Cord, Matthijs Douze, Francisco Massa, Alexandre
  Sablayrolles, and Herv{\'e} J{\'e}gou.
\newblock Training data-efficient image transformers \& distillation through
  attention.
\newblock In {\em International Conference on Machine Learning (ICML)}, 2021.

\bibitem{tsuzuku2020normalized}
Yusuke Tsuzuku, Issei Sato, and Masashi Sugiyama.
\newblock Normalized flat minima: Exploring scale invariant definition of flat
  minima for neural networks using pac-bayesian analysis.
\newblock In {\em International Conference on Machine Learning}, pages
  9636--9647. PMLR, 2020.

\bibitem{wen2018smoothout}
Wei Wen, Yandan Wang, Feng Yan, Cong Xu, Chunpeng Wu, Yiran Chen, and Hai Li.
\newblock Smoothout: Smoothing out sharp minima to improve generalization in
  deep learning.
\newblock {\em arXiv preprint arXiv:1805.07898}, 2018.

\bibitem{yun2019cutmix}
Sangdoo Yun, Dongyoon Han, Seong~Joon Oh, Sanghyuk Chun, Junsuk Choe, and
  Youngjoon Yoo.
\newblock Cutmix: Regularization strategy to train strong classifiers with
  localizable features.
\newblock In {\em Proceedings of the IEEE Conference on Computer Vision and
  Pattern Recognition (CVPR)}, 2019.

\bibitem{ZagoruykoK16}
Sergey Zagoruyko and Nikos Komodakis.
\newblock Wide residual networks.
\newblock In Richard~C. Wilson, Edwin~R. Hancock, and William A.~P. Smith,
  editors, {\em British Machine Vision Conference (BMVC)}, 2016.

\bibitem{zhang2021understanding}
Chiyuan Zhang, Samy Bengio, Moritz Hardt, Benjamin Recht, and Oriol Vinyals.
\newblock Understanding deep learning (still) requires rethinking
  generalization.
\newblock {\em Communications of the ACM}, 64(3):107--115, 2021.

\bibitem{zhang2018mixup}
Hongyi Zhang, Moustapha Cisse, Yann~N Dauphin, and David Lopez-Paz.
\newblock mixup: Beyond empirical risk minimization.
\newblock In {\em International Conference on Learning Representations (ICLR)},
  2018.

\bibitem{zhang2019lookahead}
Michael Zhang, James Lucas, Jimmy Ba, and Geoffrey~E Hinton.
\newblock Lookahead optimizer: k steps forward, 1 step back.
\newblock In {\em Advances in Neural Information Processing Systems (NeurIPS)},
  2019.

\bibitem{zhao2022penalizing}
Yang Zhao, Hao Zhang, and Xiuyuan Hu.
\newblock Penalizing gradient norm for efficiently improving generalization in
  deep learning.
\newblock In {\em International Conference on Machine Learning (ICML)}, 2022.

\bibitem{zheng2021regularizing}
Yaowei Zheng, Richong Zhang, and Yongyi Mao.
\newblock Regularizing neural networks via adversarial model perturbation.
\newblock In {\em Proceedings of the IEEE Conference on Computer Vision and
  Pattern Recognition (CVPR)}, 2021.

\bibitem{zhuang2022surrogate}
Juntang Zhuang, Boqing Gong, Liangzhe Yuan, Yin Cui, Hartwig Adam, Nicha
  Dvornek, Sekhar Tatikonda, James Duncan, and Ting Liu.
\newblock Surrogate gap minimization improves sharpness-aware training.
\newblock In {\em International Conference on Learning Representations (ICLR)},
  2022.

\end{thebibliography}
}

\newpage
\appendix

\section{Training Curves Visualization}
\begin{figure}[htbp]
 \centering
 \begin{subfigure}{0.8\linewidth}
 \centering
  \includegraphics[width=1\linewidth]{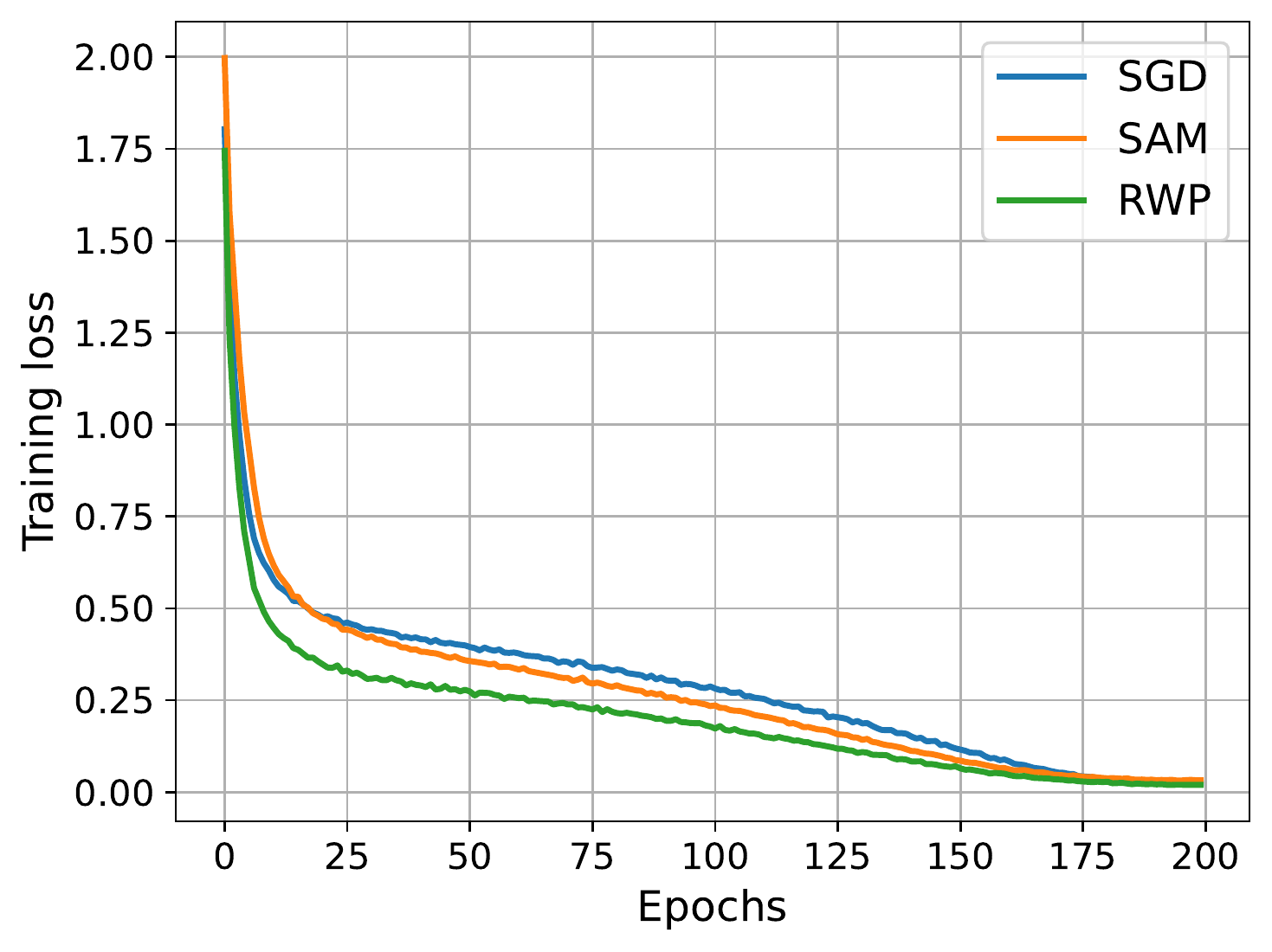}
    \caption{Training loss.}
 \end{subfigure}
 \begin{subfigure}{0.8\linewidth}
 \centering
  \includegraphics[width=1\linewidth]{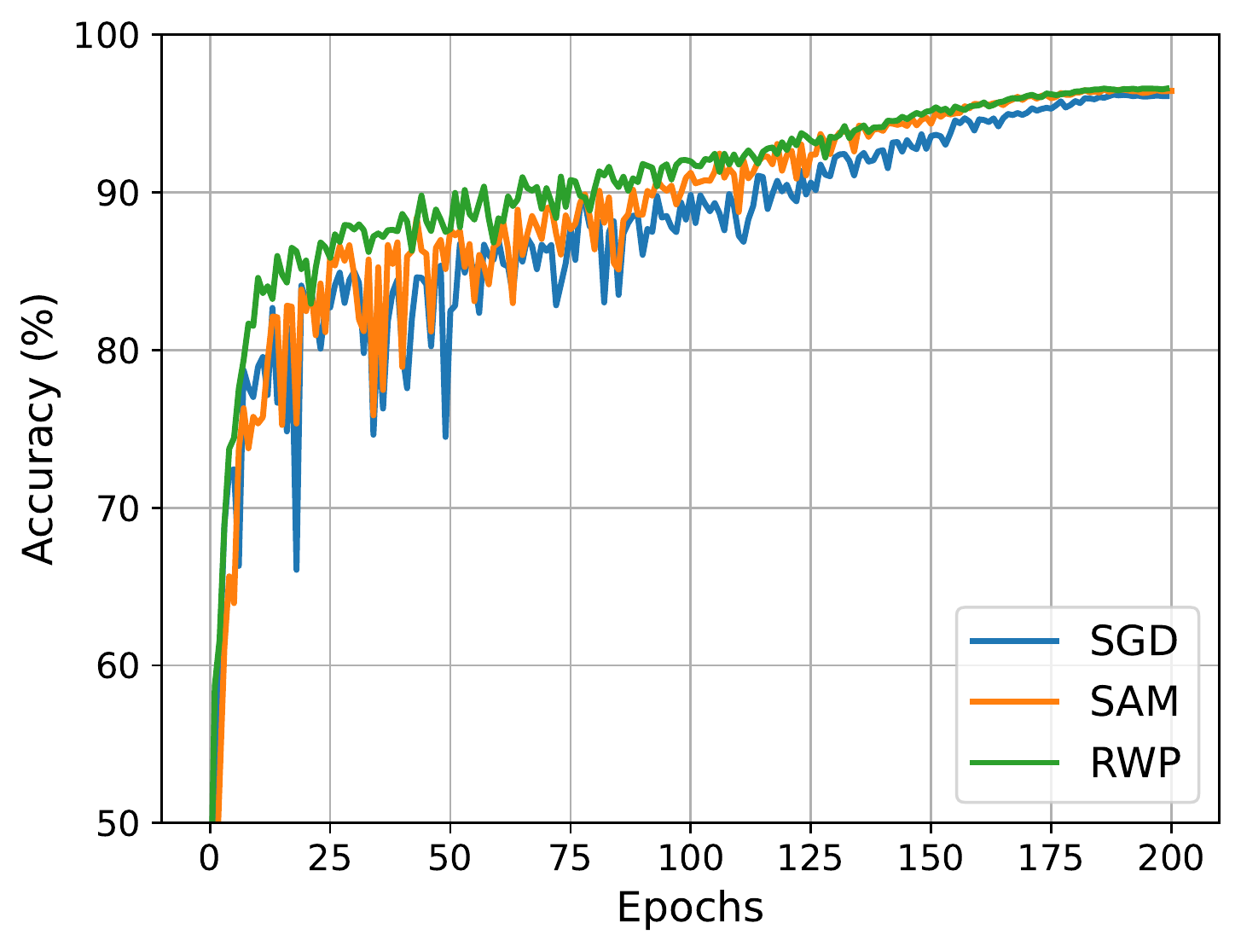}
    \caption{Test accuracy.}
 \end{subfigure}
 \caption{Training curves of ResNet-18 on CIFAR-10.}
 \label{fig:app-cifar-10}
\end{figure}

\begin{figure}[htbp]
 \centering
 \begin{subfigure}{0.8\linewidth}
 \centering
  \includegraphics[width=1\linewidth]{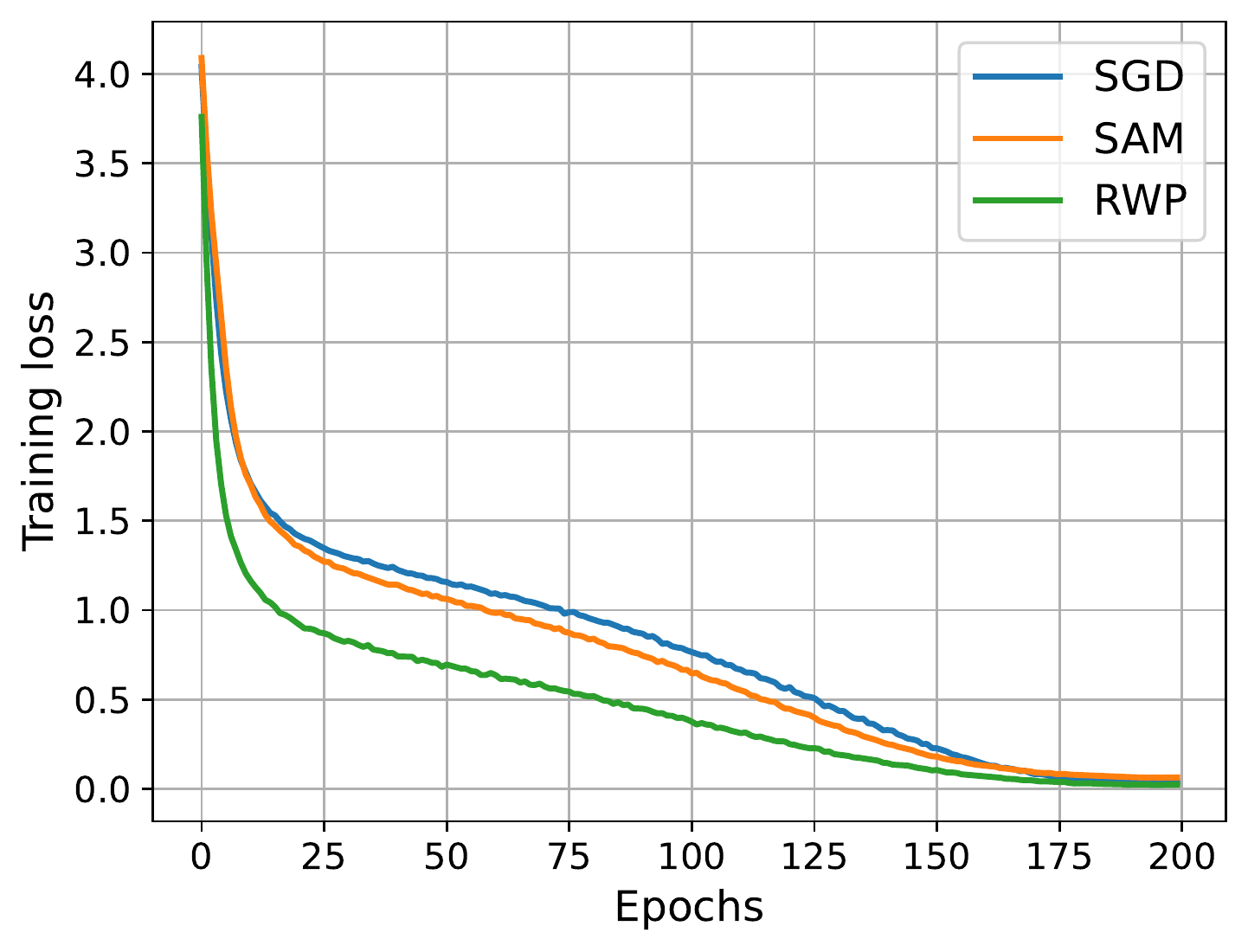}
    \caption{Training loss.}
 \end{subfigure}
 \begin{subfigure}{0.8\linewidth}
 \centering
  \includegraphics[width=1\linewidth]{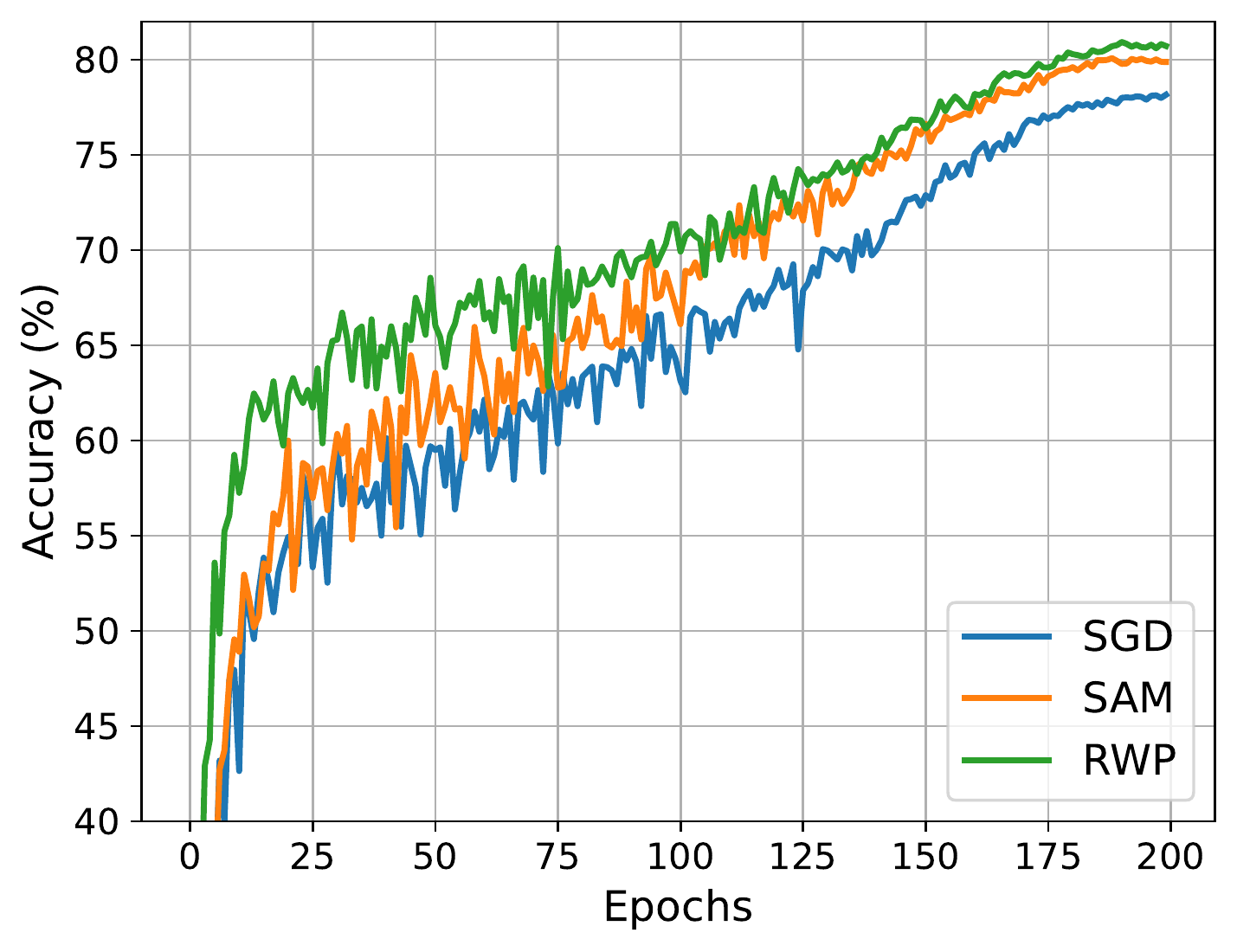}
    \caption{Test accuracy.}
 \end{subfigure}
 \caption{Training curves of ResNet-18 on CIFAR-100.}
 \label{fig:app-cifar-100}
\end{figure}

\begin{figure}[htbp]
 \centering
 \begin{subfigure}{0.8\linewidth}
 \centering
  \includegraphics[width=1\linewidth]{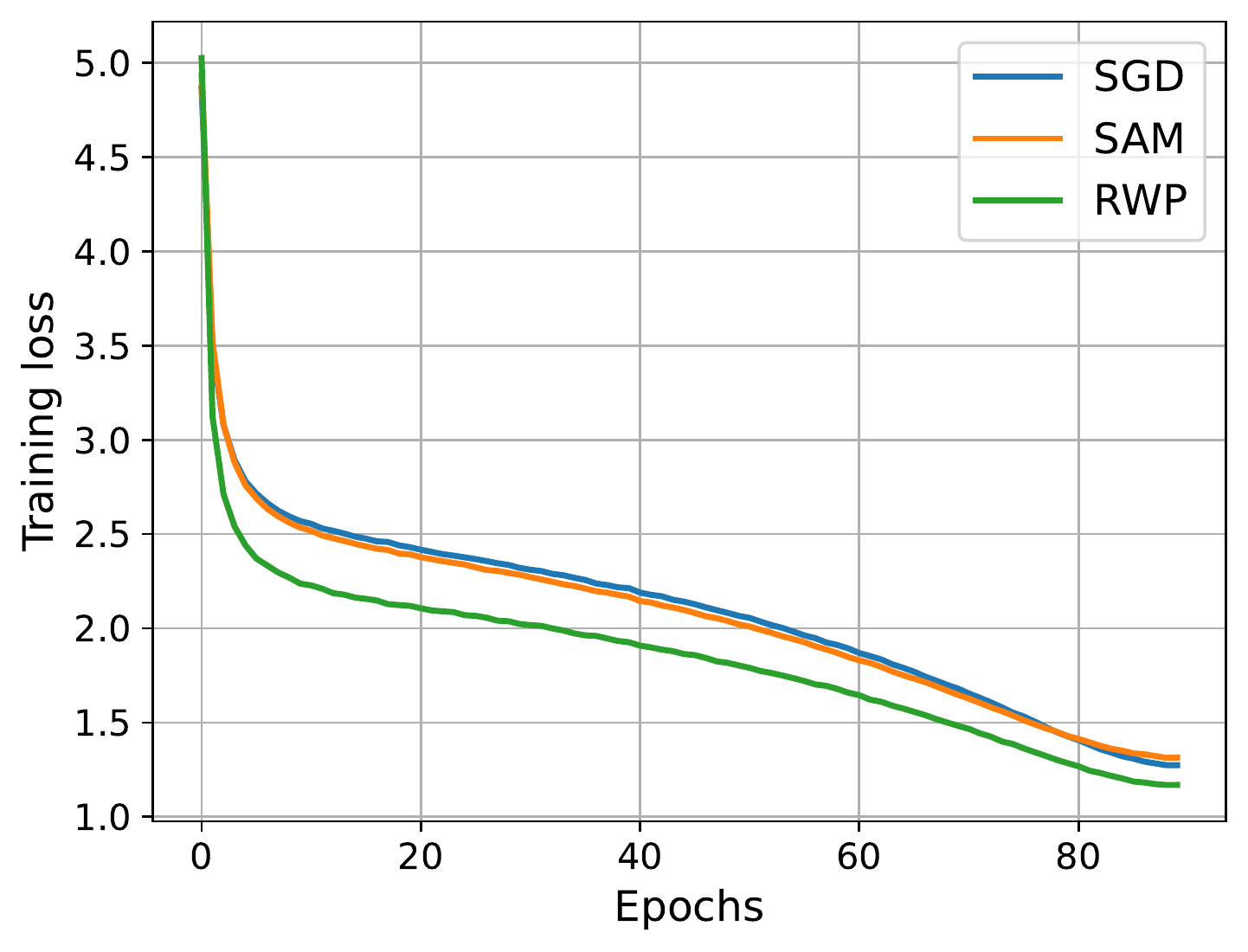}
    \caption{Training loss.}
 \end{subfigure}
 \begin{subfigure}{0.8\linewidth}
 \centering
  \includegraphics[width=1\linewidth]{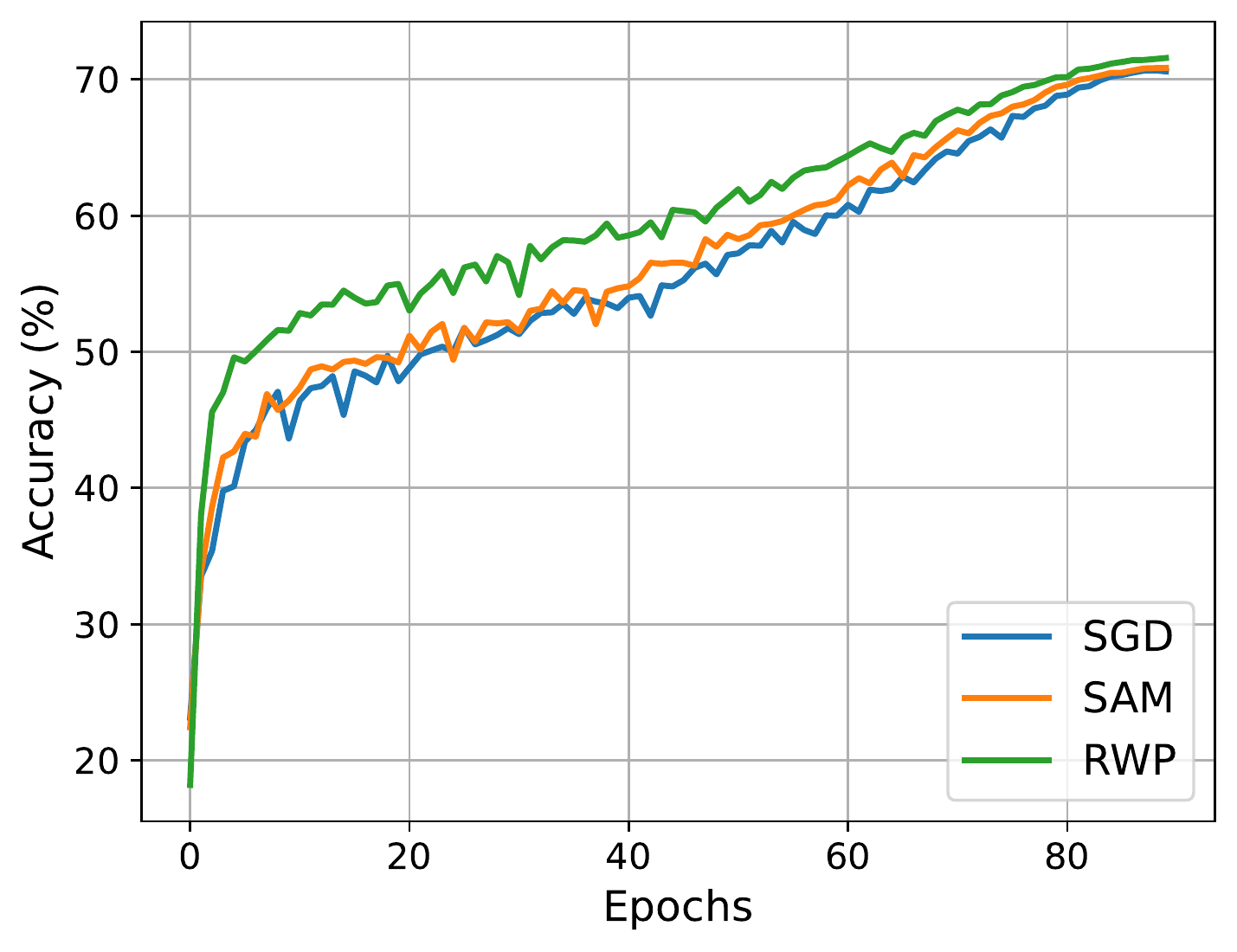}
    \caption{Test accuracy.}
 \end{subfigure}
 \caption{Training curves of ResNet-18 on ImageNet.}
 \label{fig:app-imagenet}
\end{figure}

We plot the training curves of different methods on CIFAR-10, CIFAR-100, and ImageNet datasets in \cref{fig:app-cifar-10}, \cref{fig:app-cifar-100}, \cref{fig:app-imagenet}, respectively, with ResNet-18 model. We observe that RWP could achieve faster convergence than SGD and SAM  shown in both training loss and test accuracy.


\section{Wall-clock Time Comparison}
The wall-clock time consumption of different methods is reported in \cref{tab:wall-clock}, where we use ResNet-18 model and measure the mean elapsed time for each iteration (averaged within 10 training epochs). 
We adopt the DDP module in Pytorch for a parallelized implementation of RWP.
All experiments are conducted using Nvidia Geforce GTX 2080 Ti GPU. 
We notice that RWP does not perfectly halve the training time of SAM due to the communication cost that existed across different GPUs, but is already able to remarkably reduce the training time of SAM by  $44\%-46\%$.

\begin{table}[htbp]
    \centering
    \small
    \caption{Wall-clock Time consumption for different methods.  }
    \label{tab:wall-clock}
    \begin{threeparttable}
    \begin{tabular}{l|ccc}
    \toprule
         Datasets &SGD &SAM &RWP \\
         \midrule \midrule
        CIFAR-10\tnote{*} &0.098s / iter &0.195s / iter &0.105s / iter\\ 
        CIFAR-100\tnote{*} &0.100s / iter &0.195s / iter &0.105s / iter\\ 
        ImageNet\tnote{$\dagger$} &0.277s / iter &0.537s / iter &0.303s / iter\\ 
    \bottomrule
    \end{tabular} 
    \begin{tablenotes}
        \footnotesize
        \item[*] We use 1 GPU for SGD/SAM and 2 for RWP.
        \item[{$\dagger$}] We use 2 GPUs for SGD/SAM and 4 for RWP. 
      \end{tablenotes}
    \end{threeparttable}
\end{table}

\section{Same Batch v.s. Different Batches}
In SAM and RWP, there are two gradient steps involved for each iteration, where we could adopt either the same batch data or different batches of data to calculate these two gradients. Here, we investigate the impact of this factor on the generalization performance of both methods.

We first study on RWP, where we use either the same or two different batches of data to calculate $\nabla f(\boldsymbol{w})$ and $\nabla f(\boldsymbol{w} + \bm{\epsilon}_r)$. We adopt ResNet-18 model and conduct experiments on CIFAR-10 / 100. In \cref{tab:rwp_same}, we observe that these two cases perform comparably, while RWP with different batches is slightly better. This confirms that the two gradients in RWP are independent and whether using the same batch data or not will not affect the final performance significantly. For an easy implementation with the Pytorch DDP framework, we suggest using different batches of data.

\begin{table}[htbp]
    \centering
    \caption{Comparisons on using the same / different batches for the two gradient steps in RWP with ResNet-18.}
    \label{tab:rwp_same}
    \begin{tabular}{l|ccc}
    \toprule
     Datasets   & SGD  & RWP (same)   & RWP (diff.) \\
     \midrule
     \midrule
     CIFAR-10   &96.10{\scriptsize$\pm$0.08}   &96.51{\scriptsize$\pm$0.08}      &96.68{\scriptsize$\pm$0.17}       \\
     CIFAR-100  &78.10{\scriptsize$\pm$0.39} &80.60{\scriptsize$\pm$0.10}  &80.75{\scriptsize$\pm$0.16}  \\ 
     \bottomrule
    \end{tabular}
\end{table}

Then we investigate this impact on SAM. Similarly, we will adopt either the same batch or different batches of data for calculating $\nabla f(\boldsymbol{w})$ and $\nabla f(\boldsymbol{w+\bm{\epsilon}_s})$. The results are presented in \cref{tab:sam_same}. 
We observe that SAM's performance would severely degrade to that of regular SGD training when applying different batches of data.
This indicates that the two gradient steps in SAM are crucially correlated.
Each iteration of SAM has to pay attention to a fixed batch of data. In other words, SAM tries to minimize the (expected) sharpness of loss over a randomly selected batch data instead of the sharpness over the whole datasets, which is referred to as $m$-sharpness as detailed discussed in \cite{andriushchenko2022towards}.
In this regard, the two gradient steps in SAM's iteration are never separable and $2\times$ training time is demanded.


\begin{table}[htbp]
    \centering
    \caption{Comparisons on using the same / different batches for the two gradient steps in SAM with ResNet-18.}
    \label{tab:sam_same}
    \begin{tabular}{l|ccc}
    \toprule
     Datasets   & SGD  & SAM (same)   & SAM (diff.) \\
     \midrule
     \midrule
     CIFAR-10   &96.10{\scriptsize$\pm$0.08}   &96.50{\scriptsize$\pm$0.08}      &96.16{\scriptsize$\pm$0.13}       \\
     CIFAR-100  &78.10{\scriptsize$\pm$0.39} &80.16{\scriptsize$\pm$0.25}  &78.30{\scriptsize$\pm$0.11}  \\ 
     \bottomrule
    \end{tabular}
\end{table}

\section{Results with ViT models}
\label{sec:vit}
We would like to investigate the effectiveness of our RWP on recent vision transformer models \cite{dosovitskiy2020image}. We adopt Adam \cite{kingma2014adam} as the base optimizer and train the models on CIFAR-10 / 100 from scratch with Adam, SAM, and RWP. Specifically, we select the ViT model \cite{dosovitskiy2020image} with input size 32 $\times$ 32, patch size 4, number of heads 8, and dropout rate 0.1. We train the models for 400 epochs with batch size 256, a learning rate of 0.0001, and a cosine annealing schedule.
For RWP, we set $\gamma=0.001$ and $\alpha=0.5$. 
For SAM, we perform a grid search for $\rho$ over the range $[0.00005, 0.1]$ and finally select $\rho=0.001$ for optimal.  
The results are shown in \cref{tab:vit}. 
It could be observed that SAM performs only comparably as Adam optimizer, though we have tried a lot of different choices for $\rho$. But still, our RWP has significant improvement with 1.56\% on CIFAR-10 and 1.30\% on CIFAR-100 compared to Adam. This further confirms the broad applications of RWP.



\begin{table}[htbp]
    \centering
    \caption{Comparison of Adam, SAM and RWP on ViT model.}
    \label{tab:vit}
    \begin{tabular}{l|ccc}
    \toprule
     Datasets   & Adam  & Adam-SAM   & Adam-RWP \\ 
    \midrule \midrule
     CIFAR-10   &   86.62{\scriptsize$\pm$0.10}    & 86.89{\scriptsize$\pm$0.49} & \textbf{88.18}{\scriptsize$\pm$0.44} \\
     CIFAR-100  & 63.66{\scriptsize$\pm$0.28} & 63.79{\scriptsize$\pm$0.60} & \textbf{64.96}{\scriptsize$\pm$0.74} \\ 
    \bottomrule
    \end{tabular}
\end{table}

\section{Results on Fine-tune Task}
The neural network pre-trained on the large-scale dataset usually has better transferability, that is, it can quickly and easily adapt to downstream tasks or new datasets through fine-tuning. 
To evaluate the performance of RWP on fine-tuning tasks,
we use the transformer-based model DeiT-small\cite{deit} pre-trained on ImageNet and fine-tune it on CIFAR-10 and CIFAR-100. 
Similarly, we adopt Adam \cite{kingma2014adam} as the base optimizer and compare the performance of three training methods: Adam, SAM, and RWP. We finetune the models for 20 training epochs with batch size 128, initial learning rate 0.0001, and a cosine schedule. Similar in \cref{sec:vit}, we adopt $\gamma=0.001$ and $\alpha=0.5$ for RWP, while for SAM, we search over the range $[0.00005, 0.1]$ and adopt $\rho=0.001$ for optimal.
From the experimental results in \cref{tab:fintune-app}, we observe that RWP could consistently perform better than Adam and SAM. 



\begin{table}[htbp]
    \centering
    \caption{Test accuracies (\%) for fine-tuning.}
    \label{tab:fintune-app}
    \begin{tabular}{l|ccc}
    \toprule
     Datasets   & Adam  & Adam-SAM   & Adam-RWP \\ 
    \midrule \midrule
     CIFAR-10   &   98.08{\scriptsize$\pm$0.05}    & 97.99{\scriptsize$\pm$0.10} & \textbf{98.21}{\scriptsize$\pm$0.10} \\
     CIFAR-100  & 88.64{\scriptsize$\pm$0.14} & 88.74{\scriptsize$\pm$0.13} & \textbf{89.01}{\scriptsize$\pm$0.21} \\ 
    \bottomrule
    \end{tabular}
\end{table}

\end{document}